\documentclass[10pt]{article}

\usepackage[letterpaper, margin=1in]{geometry}
\usepackage[letterpaper]{geometry}
\usepackage{algorithm}
\usepackage{algorithmic}
\usepackage{caption}
\usepackage{subcaption}
\usepackage{authblk}
\usepackage{setspace}
\usepackage{ourStyle}
\usepackage{amsmath,amssymb,graphicx,epstopdf,textpos,rotating}

\usepackage[parfill]{parskip}

\usepackage[round]{natbib}
\usepackage{color}

\newcommand{\batch}{\mathcal{B}}
\newcommand{\Var}{\mathrm{Var}}
\newcommand{\expect}{\mathbb{E}}
\allowdisplaybreaks

\title{Big Batch SGD: Automated Inference using Adaptive Batch Sizes}
\author[S. De et al.]
       {Soham De, Abhay Yadav, David Jacobs and Tom Goldstein\\
       Department of Computer Science, University of Maryland, College Park\\
       \texttt{\{sohamde, jaiabhay, djacobs, tomg\}@cs.umd.edu}}

\begin{document}
\maketitle

{\let\thefootnote\relax\footnotetext{A preliminary version of this paper appears in AISTATS 2017 (International Conference on Artificial Intelligence and Statistics), Fort Lauderdale, USA \cite{de2017automated}. This is the extended version.}}

\begin{abstract}
Classical stochastic gradient methods for optimization rely on noisy gradient approximations that become progressively less accurate as iterates approach a solution.  The large noise and small signal in the resulting gradients makes it difficult to use them for adaptive stepsize selection and automatic stopping.  We propose alternative ``big batch'' SGD schemes that adaptively grow the batch size over time to maintain a nearly constant signal-to-noise ratio in the gradient approximation. 
The resulting methods have similar convergence rates to classical SGD, and do not require convexity of the objective.  The high fidelity gradients enable automated learning rate selection and do not require stepsize decay. Big batch methods are thus easily automated and can run with little or no oversight.
\end{abstract}

\section{Introduction}

We are interested in problems of the form
\begin{equation}
\min_{x \in \mathcal{X}} \ell(x) \coloneqq \begin{cases}
\mathbb{E}_{z\sim p} [f (x ;z)], \\
 \frac{1}{N}\sum_{i=1}^N f (x ;z_i),
\end{cases}
\label{obj}
\end{equation}
where $\{z_i\}$ is a collection of data drawn from a probability distribution $p.$ We assume that $\ell$ and $f$ are differentiable, but possibly non-convex, and domain $\mathcal{X}$ is convex. In typical applications, each term $f(x;z)$ measures how well a model with parameters $x$ fits one particular data observation $z.$ The expectation over $z$ measures how well the model fits the entire corpus of data on average.  

When $N$ is large (or even infinite), it becomes intractable to exactly evaluate $\ell(x)$ or its gradient $\nabla \ell(x)$, which makes classical gradient methods impossible.  In such situations,  the method of choice for minimizing \eqref{obj} is the stochastic gradient descent (SGD) algorithm \citep{robbins1951stochastic}.  On iteration $t,$ SGD selects a batch $\batch \subset \{z_i\}$ of data uniformly at random, and then computes
\begin{samepage}
\begin{align} \label{sgd}
x_{t+1} = x_t - \alpha_t  \nabla_x \ell_\batch(x_t),\,\, \\ \text{ where } \,\,\, \ell_\batch(x)= \frac{1}{|\batch|}\sum_{z\in \batch} f (x ;z), \nonumber
\end{align}
\end{samepage}
and $\alpha_t$ denotes the stepsize used on the $t$-th iteration.
 Note that $ \expect_\batch [\nabla_x \ell_\batch(x_t)] = \nabla_x \ell(x_t),$ and so the calculated gradient $\nabla_x \ell_\batch(x_t)$ can be interpreted as a ``noisy'' approximation to the true gradient.

Because the gradient approximations are noisy, the stepsize $\alpha_t$ must vanish as $t \to \infty$ to guarantee convergence of the method.  
Typical stepsize rules require the user to find the optimal decay rate schedule, which usually requires an expensive grid search over different possible parameter values.



In this paper, we consider a ``big batch'' strategy for SGD.  Rather than letting the stepsize vanish over time as the iterates approach a minimizer, we let the minibatch $\batch$ \emph{adaptively grow} in size to maintain a constant signal-to-noise ratio of the gradient approximation. This prevents the algorithm from getting overwhelmed with noise, and guarantees convergence with an appropriate constant stepsize. Recent results \citep{keskar2016large} have shown that large \emph{fixed} batch sizes fail to find good minimizers for non-convex problems like deep neural networks. Adaptively increasing the batch size over time overcomes this limitation: intuitively, in the initial iterations, the increased stochasticity (corresponding to smaller batches) can help land the iterates near a good minimizer, and larger batches later on can increase the speed of convergence towards this minimizer.

Using this batching strategy, we show that we can keep the stepsize constant, or let it adapt using a simple Armijo backtracking line search, making the method completely adaptive with no user-defined parameters. We also derive an adaptive stepsize method based on the \citet{barzilai1988two} curvature estimate that fully automates the big batch method, while empirically enjoying a faster convergence rate than the Armijo backtracking line search.

Big batch methods that adaptively grow the batch size over time have several potential advantages over conventional small-batch SGD:
\begin{itemize}
\item Big batch methods don't require the user to choose stepsize decay parameters.   Larger batch sizes with less noise enable easy estimation of the accuracy of the approximate gradient, making it straightforward to adaptively scale up the batch size and maintain fast convergence. 
\item Backtracking line search tends to work very well when combined with big batches, making the methods completely adaptive with no parameters. A nearly constant signal-to-noise ratio also enables us to define an adaptive stepsize method based on the Barzilai-Borwein curvature estimate, that performs better empirically on a range of convex problems than the backtracking line search.
\item Higher order methods like stochastic L-BFGS typically require more work per iteration than simple SGD. When using big batches, the overhead of more complex methods like L-BFGS can be amortized over more costly gradient approximations. Furthermore, better Hessian approximations can be computed using less noisy gradient terms.
\item For a restricted class of non-convex problems (functions satisfying the Polyak-\L{}ojasiewicz Inequality), the per-iteration complexity of big batch SGD is linear and the approximate gradients vanish as the method approaches a solution, which makes it easy to define automated stopping conditions.  In contrast, small batch SGD exhibits sub-linear convergence, and the noisy gradients are not usable as a stopping criterion.
\item Big batch methods are much more efficient than conventional SGD in massively parallel/distributed settings. Bigger batches perform more computation between parameter updates, and thus allow a much higher ratio of computation to communication.
\end{itemize}

For the reasons above, big batch SGD is potentially much easier to automate and requires much less user oversight than classical small batch SGD.

\subsection{Related work}
In this paper, we focus on automating stochastic optimization methods by reducing the noise in SGD. We do this by adaptively growing the batch size to control the variance in the gradient estimates, maintaining an approximately constant signal-to-noise ratio, leading to automated methods that do not require vanishing stepsize parameters. While there has been some work on adaptive stepsize methods for stochastic optimization \citep{mahsereci2015probabilistic, schaul2013no, tan2016barzilai, kingma2014adam, zeiler2012adadelta},  the methods are largely heuristic without any kind of theoretical guarantees or convergence rates. The work in \citet{tan2016barzilai} was a first step towards provable automated stochastic methods, and we explore in this direction to show provable convergence rates for the automated big batch method.

While there has been relatively little work in provable automated stochastic methods, there has been recent interest in methods that control gradient noise. These methods mitigate the effects of vanishing stepsizes, though choosing the (constant) stepsize still requires tuning and oversight. There have been a few papers in this direction that use dynamically increasing batch sizes. In \citet{friedlander2012hybrid}, the authors propose to increase the size of the batch by a constant factor on \emph{every} iteration, and prove linear convergence in terms of the iterates of the algorithm. In \citet{byrd2012sample}, the authors propose an adaptive strategy for growing the batch size; however, the authors do not present a theoretical guarantee for this method, and instead prove linear convergence for a continuously growing batch, similar to \citet{friedlander2012hybrid}. 

Variance reduction (VR) SGD methods use an error correction term to reduce the noise in stochastic gradient estimates. The methods enjoy a provably faster convergence rate than SGD and have been shown to outperform SGD on convex problems \citep{defazio2014saga, johnson2013accelerating, schmidt2013minimizing, defazio2014finito}, as well as in parallel \citep{reddi2015variance} and distributed settings \citep{de2016efficient}. A caveat, however, is that these methods require either extra storage or full gradient computations, both limiting factors when the dataset is very large. In a recent paper \citep{harikandeh2015stopwasting}, the authors propose a growing batch strategy for a VR method that enjoys the same convergence guarantees. However, as mentioned above, choosing the constant stepsize still requires tuning. 
Another conceptually related approach is importance sampling, i.e., choosing training points such that the variance in the gradient estimates is reduced \citep{bouchard2015accelerating, csiba2016importance, needell2014stochastic}. 

\section{Big Batch SGD}

\subsection{Preliminaries and motivation}
Classical stochastic gradient methods thrive when the current iterate is far from optimal.  In this case, a small amount of data is necessary to find a descent direction, and optimization progresses efficiently.  As $x_t$ starts approaching the true solution $x^\star$, however,  noisy gradient estimates frequently fail to produce descent directions and do not reliably decrease the objective.  By choosing larger batches with less noise, we may be able to maintain descent directions on each iteration and uphold fast convergence.  This observation motivates the proposed ``big batch'' method.  We now explore this idea more rigorously.

To simplify notation, we hereon use $\nabla \ell$ to denote $\nabla_x \ell$. We wish to show that a noisy gradient approximation produces a descent direction when the noise is comparable in magnitude to the true gradient. \\

\begin{lemma}
\label{descent_lemma}
A sufficient condition for $- \nabla \ell_\mathcal{B} (x)$ to be a descent direction is
\begin{equation}
\label{sufficient_descent_cond}
 \| \nabla \ell_\mathcal{B} (x) - \nabla \ell (x) \|^2 < \| \nabla \ell_\mathcal{B} (x) \|^2. \nonumber
\end{equation}
\end{lemma}

This is a standard result in stochastic optimization (see the supplemental).  In words, if the error $\| \nabla \ell_\mathcal{B} (x) - \nabla \ell (x) \|^2$ is small relative to the gradient $\| \nabla \ell_\mathcal{B} (x) \|^2$, the stochastic approximation is a descent direction.  But how big is this error and how large does a batch need to be to guarantee this condition?  By the weak law of large numbers\footnote{We assume the random variable $\nabla f(x;z)$ is measurable and has bounded second moment.  These conditions will be guaranteed by the hypothesis of Theorem \ref{grad_bound}.}
\begin{align*}
\expect [\| \nabla \ell_\batch(x) - \nabla \ell (x) \|^2]   &= \frac{1}{|\batch|}\expect [\| \nabla f(x;z) - \nabla \ell (x) \|^2] \\&= \frac{1}{|\batch|} \Tr \Var_z \nabla f(x;z),
\end{align*}
and so we can estimate the error of a stochastic gradient if we have some knowledge of the variance of $\nabla f(x;z).$  In practice, this variance could be estimated using the sample variance of a batch  $\{\nabla f(x;z)\}_{z\in\batch}.$   However, we would like some bounds on the magnitude of this gradient to show that it is well-behaved,  and also to analyze worst-case convergence behavior.  To this end, we make the following assumption. \\

\begin{assumption}
\label{lipschitz_data}
We assume $f$ has $L_z$-Lipschitz dependence on data $z$, i.e., given two data points $z_1, z_2 \sim p(z)$, we have:
$$\|\nabla f(x;z_1)-\nabla f(x;z_2)\| \le L_z \|z_1-z_2\|.$$
\end{assumption}

Under this assumption, we can bound the error of the stochastic gradient.  The bound is uniform with respect to $x,$ which makes it rather useful in analyzing the convergence rate for big batch methods.  \\

\begin{theorem} \label{grad_bound}
Given the current iterate $x$, suppose Assumption \ref{lipschitz_data} holds and that the data distribution $p$ has bounded second moment. 
  Then the estimated gradient $\nabla \ell_\batch(x)$ has variance bounded by
  \begin{align*}
  \expect_\batch \|\nabla \ell_\batch(x)-\nabla \ell(x)\|^2 &:=  \Tr \Var_\batch(\nabla \ell_\batch(x)) \\ &\le  \frac{4L^2_z \Tr \Var_z(z)}{|\batch|},
  \end{align*}
where $z \sim p(z)$. Note the bound is uniform in $x.$
\end{theorem}

The proof is in the supplemental. Note that, using a finite number of samples, one can easily approximate the quantity $\Var_z(z)$ that appears in our bound.

\subsection{A template for big batch SGD} \label{sec:template}
 Theorem \ref{grad_bound} and Lemma \ref{descent_lemma} together suggest that we should expect $d = - \nabla \ell_\batch$ to be a descent direction reasonably often provided
\begin{align}
\label{batch_increment_condition}
\theta^2 \| \nabla \ell_\mathcal{B} (x) \|^2 &\ge \frac{1}{|\mathcal{B}|} [ \Tr \mathrm{Var}_{z} (\nabla f (x; z_i))], \\
\text{ or }\quad \theta^2 \| \nabla \ell_\mathcal{B} (x) \|^2 &\ge  \frac{4L^2_z \Tr \Var_z(z)}{|\batch|}, \nonumber
\end{align}
for some $\theta<1.$   Big batch methods capitalize on this observation. 

On each iteration $t$, starting from a point $x_t,$ the big batch method performs the following steps:
\begin{enumerate}
\item  Estimate the variance $\Tr \Var_z[\nabla f(x_t; z)],$ and a batch size $K$ large enough that
\begin{align}
\theta^2 \expect \| \nabla \ell_{\mathcal{B}_t} (x_t) \|^2 &\ge \expect  \| \nabla \ell_{\mathcal{B}_t} (x_t) - \nabla \ell (x_t) \|^2 \nonumber \\ &= \frac{1}{K}\Tr \Var_z f(x_t; z), \label{rewrite_descent_cond}
\end{align}
where $\theta \in (0,1)$ and $\batch_t$ denotes the selected batch on the $t$-th iteration with $|\batch|=K.$
\item  Choose a stepsize $\alpha_t$.
\item Perform the update $x_{t+1} = x_t - \alpha_t \nabla \ell_{\batch_t} (x_t).$
\end{enumerate}
Clearly, we have a lot of latitude in how to implement these steps using different variance estimators and different stepsize strategies.  In the following section, we show that, if condition \eqref{rewrite_descent_cond} holds, then linear convergence can be achieved using an appropriate constant stepsize.
  In subsequent sections, we address the issue of how to build practical big batch implementations using automated variance and stepsize estimators that require no user oversight.

\section{Convergence Analysis}

We now present convergence bounds for big batch SGD methods \eqref{pb_gd}. 
We study stochastic gradient updates of the form 
\begin{align}
x_{t+1} &= x_t - \alpha \nabla \ell_{\mathcal{B}_t} (x_t) = x_t - \alpha (\nabla \ell (x_t) + e_t), \label{pb_gd}
\end{align}
where $e_t = \nabla \ell_{\mathcal{B}_t} (x_t) - \nabla \ell (x_t)$, and $\mathbb{E}_{\mathcal{B}} [e_t] = 0$. Let us also define $g_t = \nabla \ell (x_t) + e_t$.

Before we present our results, we first state two assumptions about the loss function $\ell(x)$.  \\

\begin{assumption}
\label{lipschitz_assumption}
We assume that the objective function $\ell$ has $L$-Lipschitz gradients:
$$\ell(x) \leq \ell(y)+ \nabla \ell(y)^T (x - y)  + \frac{L}{2} \| x - y\|^2.$$
\end{assumption}
This is a standard smoothness assumption used widely in the optimization literature. Note that a consequence of Assumption \ref{lipschitz_assumption} is the property: $$\| \nabla \ell(x) - \nabla \ell(y) \| \le L \| x - y \|.$$

\begin{assumption}
\label{pl_inequality_assumption}
We also assume that the objective function $\ell$ satisfies the Polyak-\L{}ojasiewicz Inequality:
$$\| \nabla \ell(x) \|^2 \ge 2\mu (\ell(x) - \ell(x\opt)).$$
\end{assumption}
Note that this inequality does \emph{not} require $\ell$ to be convex, and is, in fact, a weaker assumption than what is usually used. It does, however, imply that every stationary point is a global minimizer \citep{karimi2016linear, polyak1963gradient}.

We now present a result that establishes an upper bound on the objective value in terms of the error in the gradient of the sampled batch. We present all the proofs in the Supplementary Material.  \\

\begin{lemma}
\label{objective_error}
Suppose we apply an update of the form  \eqref{pb_gd}  where the batch $\batch_t$ is uniformly sampled from the distribution $p$ on each iteration $t$. If the objective $\ell$ satisfies Assumption  \ref{lipschitz_assumption},  then we have
\begin{align*}\small
\mathbb{E} [\ell(x_{t+1}) - \ell(x\opt)] \le  \expect \big[\ell (x_t) - \ell(x\opt)  - \big (\alpha - \frac{L \alpha^2}{2} \big) \| \nabla \ell(x_t) \|^2  + \frac{L \alpha^2}{2} \| e_t \|^2 \big].
\end{align*}
Further, if the objective $\ell$ satisfies the PL Inequality (Assumption \ref{pl_inequality_assumption}), we have:
\begin{align*}
\mathbb{E} [\ell(x_{t+1}) - \ell(x\opt)] \le \Big(1 - 2\mu \big (\alpha - \frac{L \alpha^2}{2} \big) \Big) \expect [\ell (x_t) - \ell(x\opt)] + \frac{L \alpha^2}{2}  \mathbb{E}  \| e_t \|^2.
\end{align*}
\end{lemma}
Using Lemma \ref{objective_error}, we now provide convergence rates for big batch SGD.  \\
\begin{theorem} \label{strong_rate}
Suppose $\ell$ satisfies Assumptions \ref{lipschitz_assumption} and \ref{pl_inequality_assumption}.  Suppose further that on each iteration the batch size is large enough to satisfy \eqref{rewrite_descent_cond} for $\theta\in (0,1)$.  
 If $0 \le\alpha<\frac{2}{L\beta},$ where $\beta = \frac{\theta^2 + (1-\theta)^2}{(1-\theta)^2},$ then we get the following linear convergence bound for big batch SGD using updates of the form \ref{pb_gd}:
\aln{
\mathbb{E} [\ell(x_{t+1}) - \ell(x\opt)] \le \gamma \cdot \expect [\ell (x_t) - \ell(x\opt)], \nonumber
}
where $\gamma = \big(1 - 2\mu  (\alpha - \frac{L \alpha^2\beta}{2} ) \big)$. Choosing the optimal stepsize of $\alpha = \frac{1}{\beta L}$, we get
\aln{
\mathbb{E} [\ell(x_{t+1}) - \ell(x\opt)] \le \big( 1 - \frac{\mu}{\beta L}  \big) \cdot \expect [\ell (x_t) - \ell(x\opt)]. \nonumber
}
\end{theorem}
Note that the above linear convergence rate bound holds without requiring convexity. Comparing it with the convergence rate of deterministic gradient descent under similar assumptions, we see that big batch SGD suffers a slowdown by a factor $\beta$, due to the noise in the estimation of the gradients. We now present a result proving a $\mathcal{O}(1/t)$ convergence rate for general smooth convex functions. \\

\begin{theorem}
\label{sublinear}
Suppose $\ell$ satisfies Assumptions \ref{lipschitz_assumption}, is convex, and condition \eqref{rewrite_descent_cond} is satisfied on each iteration. Then we get sub-linear convergence of the form:
\begin{equation}
\mathbb{E} [\ell(x_{t}) - \ell(x\opt)] \le \frac{ \| x_{0} - x\opt \|^2 }{(2\alpha - 2L \alpha^2 \beta) (t+1)}  = \mathcal{O} (1/t),\nonumber
\end{equation}
where $\beta = \frac{\theta^2 + (1-\theta)^2}{(1-\theta)^2}$ and $\alpha < \frac{1}{L\beta}$. Choosing the optimal step size of $\alpha = \frac{1}{2L\beta}$, we get
\begin{equation}
\mathbb{E} [\ell(x_{t}) - \ell(x\opt)] \le \frac{ 2L\beta \| x_{0} - x\opt \|^2 }{t+1}  = \mathcal{O} (1/t).\nonumber
\end{equation}
\end{theorem}

\subsection{Comparison to classical SGD} \label{sec:theory_compare}
Conventional small batch SGD methods can attain only $O(1/t)$ convergence for strongly convex problems, thus requiring $O(1/\epsilon)$ gradient evaluations to achieve an optimality gap less than $\epsilon$, and this has been shown to be {\em optimal} in the online setting (i.e., the infinite data setting) \citep{rakhlin2011making}. In the previous section, however, we have shown that big batch SGD methods converge linearly in the number of iterations, under a weaker assumption than strong convexity, in the online setting. Unfortunately, per-iteration convergence rates are not a fair comparison between these methods because the cost of a big batch iteration grows with the iteration count, unlike classical SGD.  For this reason, it is interesting to study the convergence rate of big batch SGD as a function of {\em gradient evaluations}.  
   
   From Lemma \ref{objective_error}, we see that we should not expect to achieve an optimality gap less than $\epsilon$ until we have: $ \frac{L \alpha^2}{2}  \mathbb{E}_{\batch_t}  \| e_t \|^2 <\epsilon.$  In the worst case, by Theorem \ref{grad_bound}, this requires 
   $\frac{L \alpha^2}{2} \frac{4L^2_z \Tr \Var_z(z)}{|\batch|} <\epsilon,$ or $|\batch|  \ge O(1/\epsilon)$ gradient evaluations. Note that in the online or infinite data case, this is an optimal bound, and matches that of other SGD methods. 
   
   We choose to study the infinite sample case since the finite sample case is fairly trivial with a growing batch size: asymptotically, the batch size becomes the whole dataset, at which point we get the same asymptotic behavior as deterministic gradient descent, achieving linear convergence rates. 

\section{Practical Implementation with Backtracking Line Search}

While one could implement a big batch method using analytical bounds on the gradient and its variance (such as that provided by Theorem \ref{grad_bound}),  the purpose of big batch methods is to enable automated adaptive estimation of algorithm parameters.  Furthermore, the stepsize bounds provided by our convergence analysis, like the stepsize bounds for classical SGD, are fairly conservative and more aggressive stepsize choices are likely to be more effective.  

The framework outlined in Section \ref{sec:template} requires two ingredients:  estimating the batch size and estimating the stepsize.   Estimating the batch size needed to achieve \eqref{rewrite_descent_cond} is fairly straightforward.  We start with an initial batch size $K,$ and draw a random batch $\batch$ with $|\batch|=K.$  We then compute the stochastic gradient estimate $\nabla \ell_\batch(x_t)$
 and the sample variance 
  \aln{ \label{varest}
   V_\batch  &:=   \frac{1}{|\batch|-1}\sum_{z\in \batch} \| \nabla f(x_t;z) -\nabla \ell_\batch(x_t) \|^2 \nonumber \\
   &\approx    \Tr \mathrm{Var}_{z\in \batch} (\nabla f (x_t; z)).
  } 
  We then test whether $\|\nabla \ell_\batch(x_t)\|^2 > V_\batch/|\batch|$ as a proxy for \eqref{rewrite_descent_cond}. If this condition holds, we proceed with a gradient step, else we increase the batch size $K\gets K+\delta_K,$ and check our condition again. 
 We fix $\delta_K= 0.1K$ for all our experiments. 
  Our aggressive implementation also simply chooses
   $\theta = 1$.  The fixed stepsize big batch method is listed in Algorithm \ref{prog_batching}.
  
   \begin{algorithm}[t]
  \begin{algorithmic}[1]
    \STATE \textbf{initialize} starting pt. $x_0$, stepsize $\alpha,$ initial batch size $K>1$, batch size increment $\delta_k$
    \WHILE {not converged}
    \STATE Draw random batch with size $|\batch|=K$
        \STATE  Calculate $V_\batch$ and $\nabla \ell_\batch(x_t)$ using \eqref{varest}
    \WHILE {$ \|\nabla \ell_\batch(x_t)\|^2 \le V_\batch/K$}
    \STATE  Increase batch size $K\gets K+\delta_K$
    \STATE Sample more gradients
    \STATE   Update $V_\batch$ and  $\nabla \ell_\batch(x_t)$
    \ENDWHILE
        \STATE   $x_{t+1} = x_t - \alpha \nabla \ell_\batch(x_t)$
            \ENDWHILE
  \end{algorithmic} 
  \caption{Big batch SGD: fixed stepsize}  
  \label{prog_batching}
\end{algorithm}

  We also consider a backtracking variant of SGD that adaptively tunes the stepsize.  This method selects batch sizes using the same criterion \eqref{varest} as in the constant stepsize case.  However, after a batch has been selected, a backtracking Armijo line search is used to select a stepsize.  In the Armijo line search, we keep decreasing the stepsize by a constant factor (in our case, by a factor of 2) until the following condition is satisfied on each iteration:
\begin{align}
\ell_\batch (x_{t+1}) \le \ell_\batch (x_t) - c \alpha_t \| \nabla \ell_\batch (x_t) \|^2, \label{armijo_cond}
\end{align}
  where $c$ is a parameter of the line search usually set to $0 < c \le 0.5$.
 We now present a convergence result of big batch SGD using the Armijo line search.  \\

\begin{theorem} \label{backtracking}
Suppose that $\ell$ satisfies Assumptions \ref{lipschitz_assumption} and \ref{pl_inequality_assumption} and on each iteration, and the batch size is large enough to satisfy \eqref{rewrite_descent_cond} for $\theta\in (0,1)$.
If an Armijo line search, given by \eqref{armijo_cond}, is used, and the stepsize is decreased by a factor of 2 failing \eqref{armijo_cond}, then we get the following linear convergence bound for big batch SGD using updates of the form \ref{pb_gd}:
\begin{align*}
\mathbb{E} [\ell(x_{t+1}) - \ell(x\opt)] \le \gamma \cdot \mathbb{E} [\ell (x_t) - \ell(x\opt)],
\end{align*}
where $\gamma = \Big( 1 - 2c\mu \min \big( \alpha_0, \frac{1}{2\beta L} \big) \Big)$ and  $0 < c \le 0.5$.
If the initial stepsize $\alpha_0$ is set large enough such that $\alpha_0 \ge \frac{1}{2\beta L}$, then we get:
\begin{align*}
\mathbb{E} [\ell(x_{t+1}) - \ell(x\opt)] \le \Big(1 - \frac{c\mu}{\beta L} \Big) \mathbb{E} [\ell (x_t) - \ell(x\opt)].
\end{align*}
\end{theorem}
In practice, on iterations where the batch size increases, we double the stepsize before running line search to prevent the stepsizes from decreasing monotonically. The complete details are listed in Algorithm \ref{prog_batching_armijo}.

\begin{algorithm}[ht]
  \begin{algorithmic}[1]
    \STATE \textbf{initialize} starting pt. $x_0$, initial stepsize $\alpha,$ initial batch size $K>1$, batch size increment $\delta_k$, backtracking line search parameter $c$, flag $F = 0$
    \WHILE {not converged}
    \STATE Draw random batch with size $|\batch|=K$
        \STATE  Calculate $V_\batch$ and $\nabla \ell_\batch(x_t)$ using \eqref{varest}
    \WHILE {$ \|\nabla \ell_\batch(x_t)\|^2 \le V_\batch/K$}
    \STATE  Increase batch size $K\gets K+\delta_K$
    \STATE Sample more gradients
    \STATE   Update $V_\batch$ and  $\nabla \ell_\batch(x_t)$
    \STATE  Set flag $F = 1$
    \ENDWHILE
    \IF  {flag $F == 1$}
    \STATE $\alpha \gets \alpha * 2$
    \STATE Reset flag $F = 0$
    \ENDIF
    \WHILE {$\ell_\batch (x_t - \alpha \nabla \ell_\batch(x_t)) > \ell_\batch (x_t) - c \alpha_t \| \nabla \ell_\batch (x_t) \|^2$}
    \STATE $\alpha \gets \alpha/2$
    \ENDWHILE
        \STATE   $x_{t+1} = x_t - \alpha \nabla \ell_\batch(x_t)$
            \ENDWHILE
  \end{algorithmic} 
  \caption{Big batch SGD: backtracking line search}  
  \label{prog_batching_armijo}
\end{algorithm}

\section{Adaptive Step Sizes using the Barzilai-Borwein Estimate}
\label{sec:adaptive_bb}

While the Armijo backtracking line search leads to an automated big batch method, the stepsize sequence is monotonic (neglecting the heuristic mentioned in the previous section). In this section, we derive a non-monotonic stepsize scheme that uses curvature estimates to propose new stepsize choices.  

Our derivation follows the classical adaptive \citet{barzilai1988two} (BB) method. The BB methods fits a quadratic model to the objective on each iteration, and a stepsize is proposed that is optimal for the local quadratic model \citep{GoldsteinStuderBaraniuk:2014}.
To derive the analog of the BB method for stochastic problems, we consider quadratic approximations of the form $\ell(x) = \expect_\phi f(x; \phi)$, where $f(x; \phi) = \frac{\nu}{2} \| x-\phi \|^2$ and  $\phi \sim \mathcal{N}(x\opt, \sigma^2I)$. We now derive the optimal stepsize on each iteration for this quadratic approximation (for complete details of this derivation see the supplemental).
We can write:
\begin{align*}
\ell (x) = \expect_\phi f(x; \phi) = \frac{\nu}{2} \big( \| x - x\opt \|^2 + d\sigma^2 \big).
\end{align*}
Further, notice that:
\begin{align*}
\expect_\phi [\nabla \ell(x) ] &= \nu (x - x\opt), \quad \text{and} \\
\Tr \Var_\phi [\nabla \ell(x) ] &= d \nu^2 \sigma^2.
\end{align*}
Now, we can rewrite the big batch SGD update as:
\begin{align*}
x_{t+1} &= x_t - \alpha_t  \frac{1}{|\batch|}\sum_{i \in \batch} \nu (x_t - \phi_i) \\
&= (1-\nu \alpha_t) x_t + \nu \alpha_t x\opt + \frac{\nu \sigma \alpha_t}{|\batch|} \sum_{i \in\batch} \xi_i,
\end{align*}
where we write $\phi_i = x\opt + \sigma \xi_i$ with $\xi_i \sim \mathcal{N}(0,1)$. Thus, the expected value of the function is:
\begin{align*}
\expect [\ell(x_{t+1}) ]  
=  \frac{\nu}{2} \big( \| (1-\nu \alpha_t) (x_t - x\opt) \|^2  +  (1 + \frac{\nu^2 \alpha_t^2}{|\batch|}) d\sigma^2  \big).
\end{align*} 
Minimizing $\expect [\ell(x_{t+1}) ]$ w.r.t. $\alpha_t$ we get:
\begin{align}
\alpha_t 
&= \frac{1}{\nu} \cdot \frac{ \expect \big\|  \nabla \ell_{\batch_t} (x_t)  \big\|^2  - \frac{1}{|\batch_t|}  \Tr \Var [ \nabla f (x_t) ] }{ \expect \big\| \nabla \ell_{\batch_t} (x_t)   \big\|^2  } \nonumber \\
&= \frac{1}{\nu} \cdot \bigg( 1 -  \frac{  \frac{1}{|\batch_t|}  \Tr \Var [ \nabla f (x_t) ] }{ \expect \big\| \nabla \ell_{\batch_t} (x_t)   \big\|^2  }  \bigg) \label{bb_step}.
\end{align}
Here $\nu$ denotes the curvature of the quadratic approximation. Note that, in the case of {\em deterministic} gradient descent, the optimal stepsize is simply $1/\nu$ \citep{GoldsteinStuderBaraniuk:2014}.

We estimate the curvature $\nu_t$ on each iteration using the BB least-squares rule \citep{barzilai1988two}:
\begin{align}
\nu_t = \frac{  \langle x_t - x_{t-1},  \nabla \ell_{\batch_t} (x_t) - \nabla \ell_{\batch_{t}} (x_{t-1}) \rangle  }{ \| x_t - x_{t-1} \|^2  }. \label{curvature}
\end{align}
Thus, each time we sample a batch $\batch_t$ on the $t$-th iteration, we calculate the gradient on that batch in the previous iterate, i.e., we calculate $\nabla \ell_{\batch_t} (x_{t-1})$. This gives us an approximate curvature estimate, with which we derive the stepsize $\alpha_t$ using \eqref{bb_step}.

\subsection{Convergence Proof}
Here we prove convergence for the adaptive stepsize method described above. For the convergence proof, we first state two assumptions: \\

\begin{assumption}
\label{stronger_lipschitz_assumption}
Each $f$ has $L$-Lipschitz gradients:
$$f(x) \leq f(y)+ \nabla f(y)^T (x - y)  + \frac{L}{2} \| x - y\|^2. $$
\end{assumption}

\begin{assumption}
\label{rsi}
Each $f$ is $\mu$-strongly convex:
$$\langle \nabla f(x) - \nabla f(y), x - y \rangle \ge \mu \| x - y \|^2.$$
\end{assumption}
Note that both assumptions are stronger than Assumptions \ref{lipschitz_assumption} and \ref{pl_inequality_assumption}, i.e., Assumption \ref{stronger_lipschitz_assumption} implies \ref{lipschitz_assumption} and Assumption \ref{rsi} implies \ref{pl_inequality_assumption} \citep{karimi2016linear}. Both are very standard assumptions frequently used in the convex optimization literature.

Also note that from \eqref{bb_step}, we can lower bound the stepsize as:
$$\alpha_t \ge (1 - \theta^2)/\nu.$$
Thus, the stepsize for big batch SGD is scaled down by \emph{at most} $1-\theta^2$. 
For simplicity, we assume that the stepsize is set to this lower bound: $\alpha_t = (1-\theta^2)/\nu_t$. Thus, from Assumptions \ref{stronger_lipschitz_assumption} and \ref{rsi}, we can bound $\nu_t$, and also $\alpha_t$, as follows:
\begin{align*}
\mu \le \nu_t \le L \quad \implies \quad \frac{1- \theta^2}{L} \le \alpha_t \le \frac{1-\theta^2}{\mu}.
\end{align*}

From Theorem \ref{strong_rate}, we see that we have linear convergence with the adaptive stepsize method when:
\begin{align*}
1 - 2\mu \big (\alpha - \frac{L \alpha^2\beta}{2} \big) &\le 1 - \frac{2 (1 - \theta^2)}{\kappa} + \beta (1 - \theta^2)^2 \kappa < 1, \\
\implies \kappa^2 &< \frac{2}{\beta (1 - \theta^2)},
\end{align*}
where $\kappa = L/\mu$ is the condition number. We see that the adaptive stepsize method enjoys a linear convergence rate when the problem is well-conditioned. In the next section, we talk about ways to deal with poorly-conditioned problems.

\subsection{Practical Implementation}
To achieve robustness of the algorithm for poorly conditioned problems, we include a backtracking line search after calculating \eqref{bb_step}, to ensure that the stepsizes do not blow up. Further, instead of calculating two gradients on each iteration ($\nabla \ell_{\batch_t} (x_t)$ and  $\nabla \ell_{\batch_t} (x_{t-1})$), our implementation uses the same batch (and stepsize) on two consecutive iterations. Thus, one parameter update takes place for each gradient calculation.

We found the stepsize calculated from \eqref{bb_step} to be noisy when the batch is small. While this did not affect long-term performance, we perform a smoothing operation to even out the stepsizes and make performance more predictable. Let $\tilde \alpha_t$ denote the stepsize calculated from \eqref{bb_step}. Then, the stepsize on each iteration is given by 
$$ \alpha_t = \Big(  1 - \frac{|\batch|}{N} \Big) \alpha_{t-1}  + \frac{ |\batch|}{N}  \tilde \alpha_{t}. $$
This ensures that the update is proportional to how accurate the estimate on each iteration is. This simple smoothing operation seemed to work very well in practice as shown in the experimental section. Note that when $| \batch_t | = N$, we just use $\alpha_t = 1/\nu_t$. Since there is no noise in the algorithm in this case, we use the optimal stepsize for a deterministic algorithm. Algorithm \ref{prog_batching_bb} shows the complete details.

 \begin{algorithm}[H]
  \begin{algorithmic}[1]
    \STATE \textbf{initialize} starting pt. $x$, initial stepsize $\alpha,$ initial batch size $K>1$, batch size increment $\delta_k$, backtracking line search parameter $c$
    \WHILE {not converged}
    \STATE Draw random batch with size $|\batch|=K$
        \STATE  Calculate $V_\batch$ and $G_\batch = \nabla \ell_\batch(x)$ using \eqref{varest}
    \WHILE {$ \|G_\batch\|^2 \le V_\batch/K$}
    \STATE  Increase batch size $K\gets K+\delta_K$
    \STATE Sample more gradients
    \STATE   Update $V_\batch$ and  $G_\batch$
    \ENDWHILE
    \WHILE {$\ell_\batch (x - \alpha \nabla \ell_\batch(x)) > \ell_\batch (x) - c \alpha \| \nabla \ell_\batch (x) \|^2$}
    \STATE $\alpha \gets \alpha/2$
    \ENDWHILE
        \STATE   $x \gets x - \alpha \nabla \ell_\batch(x)$
        \IF {$K < N$}
   \STATE  Calculate $\tilde\alpha = (1 - V_B/(K \|G_\batch\|^2))/\nu $ using \eqref{bb_step} and \eqref{curvature}
   \ELSE
   \STATE Calculate $\tilde \alpha = 1/\nu$ using \eqref{curvature}
        \ENDIF
      \STATE Stepsize smoothing: $\alpha \gets \alpha (1 - K/N)   + \tilde \alpha K/N  $
    \WHILE {$\ell_\batch (x - \alpha \nabla \ell_\batch(x)) > \ell_\batch (x) - c \alpha \| \nabla \ell_\batch (x) \|^2$}
    \STATE $\alpha \gets \alpha/2$
    \ENDWHILE
       \STATE   $x \gets x - \alpha \nabla \ell_\batch(x)$
            \ENDWHILE
  \end{algorithmic} 
  \caption{Big batch SGD: with BB stepsizes}  
  \label{prog_batching_bb}
\end{algorithm}

\begin{figure*}[!t]
  \centering
  \begin{subfigure}[t]{0.32\textwidth}
    \includegraphics[width=\textwidth]{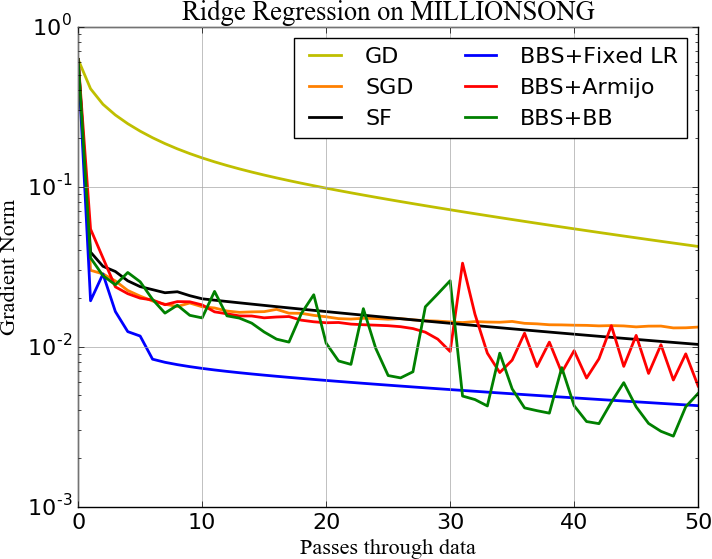}
  \end{subfigure}
  \hfill
  \begin{subfigure}[t]{0.32\textwidth}
    \includegraphics[width=\textwidth]{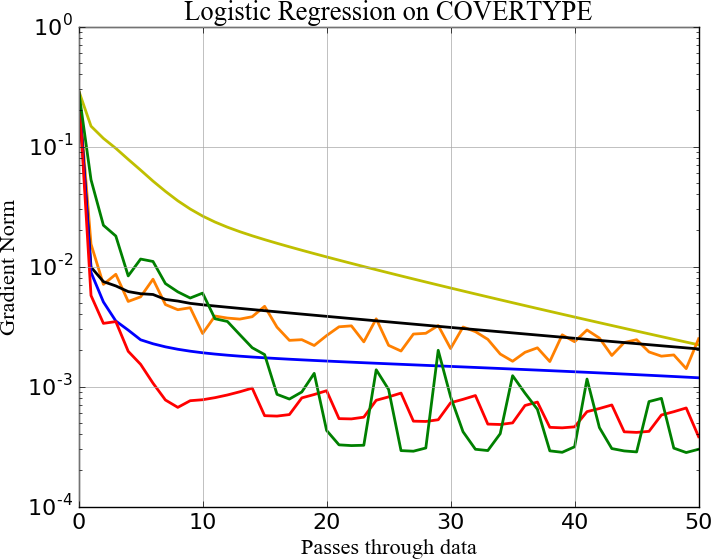}
  \end{subfigure}
  \hfill
  \begin{subfigure}[t]{0.32\textwidth}
    \includegraphics[width=\textwidth]{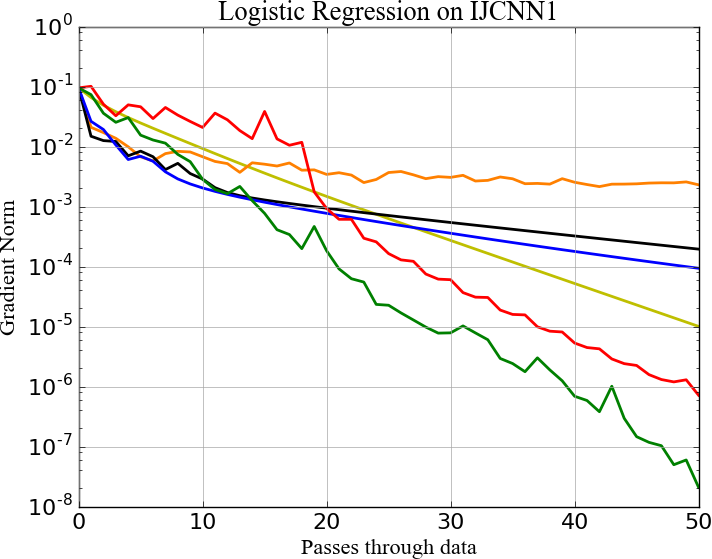}
  \end{subfigure}
\hfill
  \begin{subfigure}[t]{0.32\textwidth}
    \includegraphics[width=\textwidth]{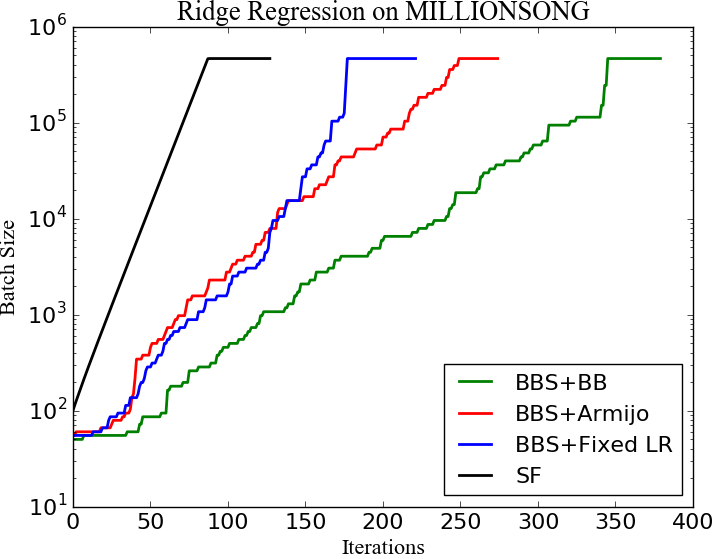}
  \end{subfigure}
  \hfill
  \begin{subfigure}[t]{0.32\textwidth}
    \includegraphics[width=\textwidth]{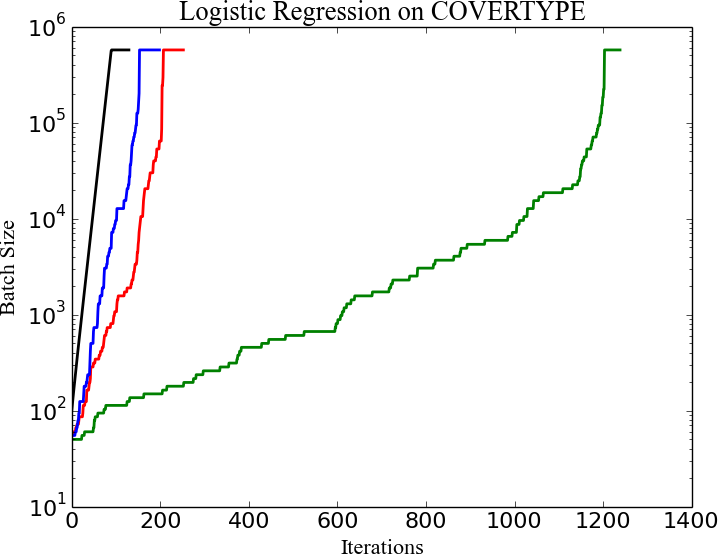}
  \end{subfigure}
  \hfill
  \begin{subfigure}[t]{0.32\textwidth}
    \includegraphics[width=\textwidth]{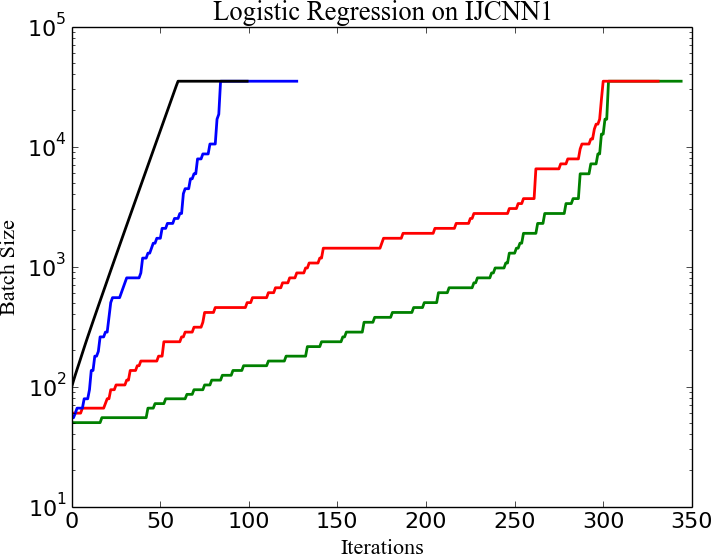}
  \end{subfigure}
  \hfill
  \begin{subfigure}[t]{0.32\textwidth}
    \includegraphics[width=\textwidth]{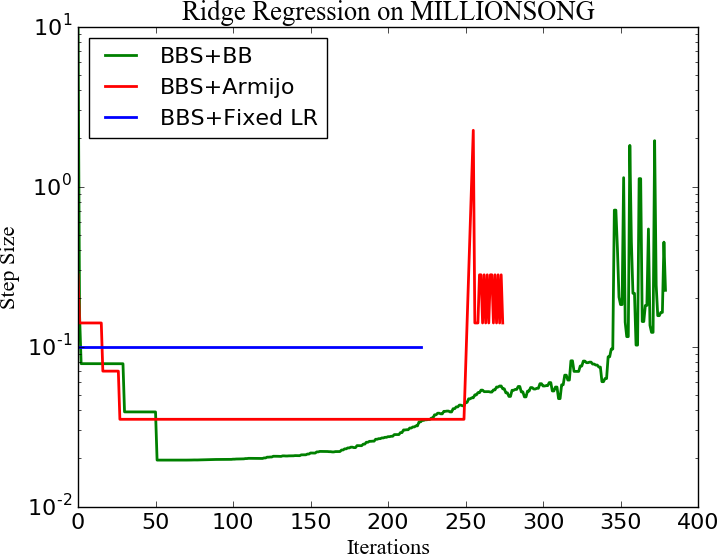}
  \end{subfigure}
  \hfill
  \begin{subfigure}[t]{0.32\textwidth}
    \includegraphics[width=\textwidth]{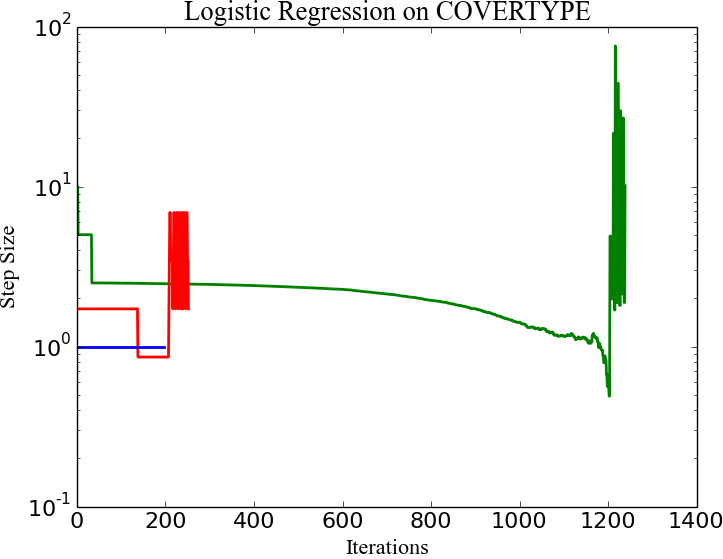}
  \end{subfigure}
  \hfill
  \begin{subfigure}[t]{0.32\textwidth}
    \includegraphics[width=\textwidth]{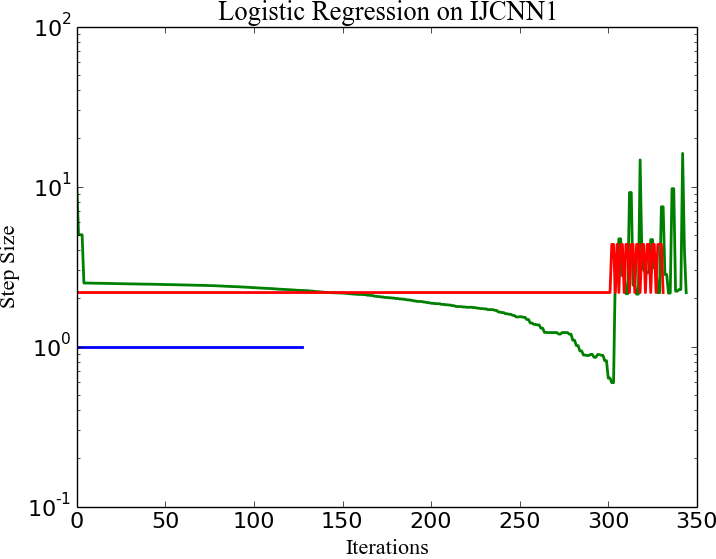}
  \end{subfigure}
  \caption{Convex experiments. Left to right: Ridge regression on MILLIONSONG; Logistic regression on COVERTYPE; Logistic regression on IJCNN1. The top row shows how the norm of the true gradient decreases with the number of epochs, the middle and bottom rows show the batch sizes and stepsizes used on each iteration by the big batch methods. Here `passes through the data' indicates number of epochs, while `iterations' refers to the number of parameter updates used by the method (there may be multiple iterations during one epoch).}
  \label{fig:convex}
\end{figure*}

\begin{figure*}[t]
  \centering
  \begin{subfigure}[t]{0.32\textwidth}
    \includegraphics[width=\textwidth]{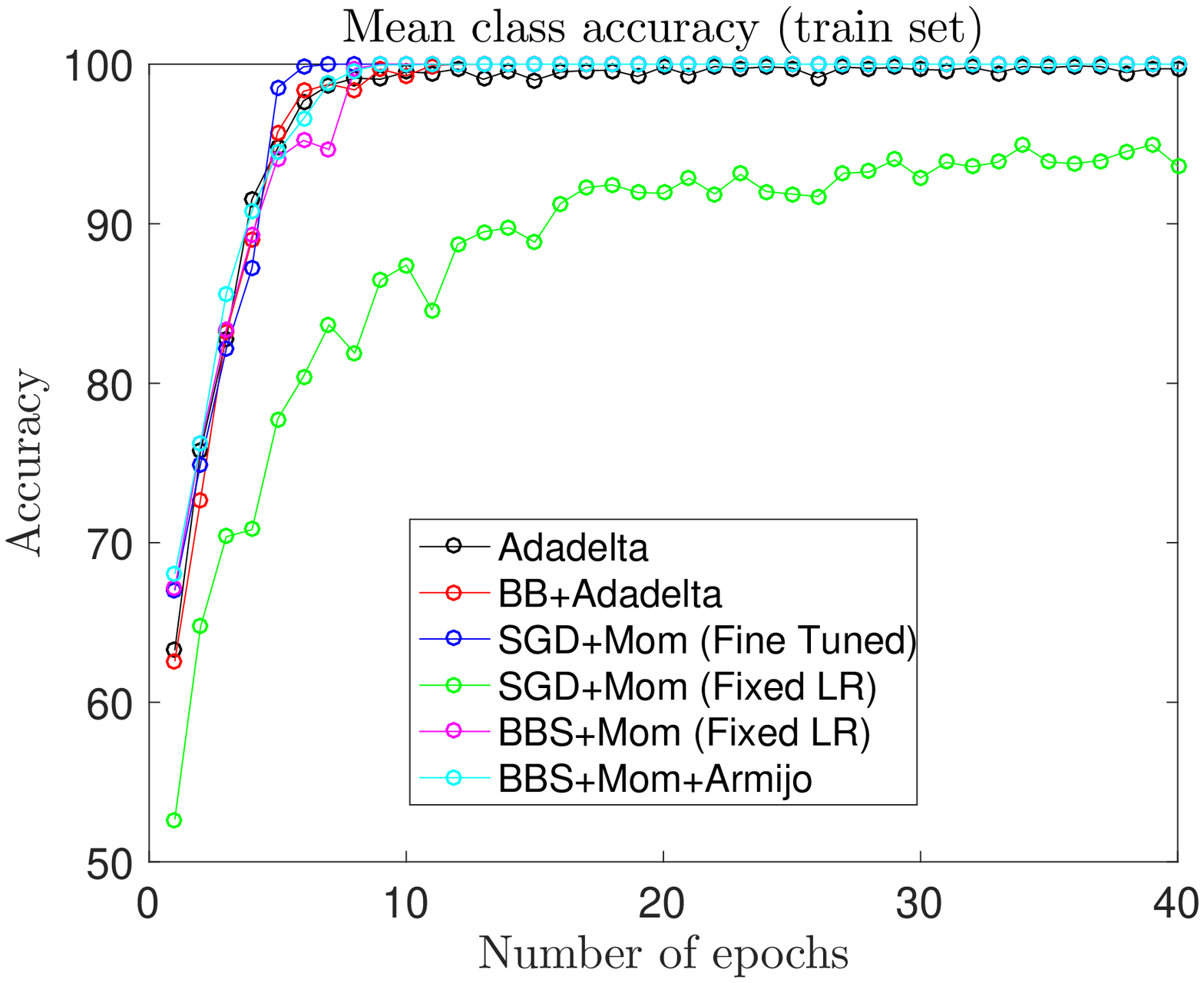}
  \end{subfigure}
  \hfill
  \begin{subfigure}[t]{0.32\textwidth}
    \includegraphics[width=\textwidth]{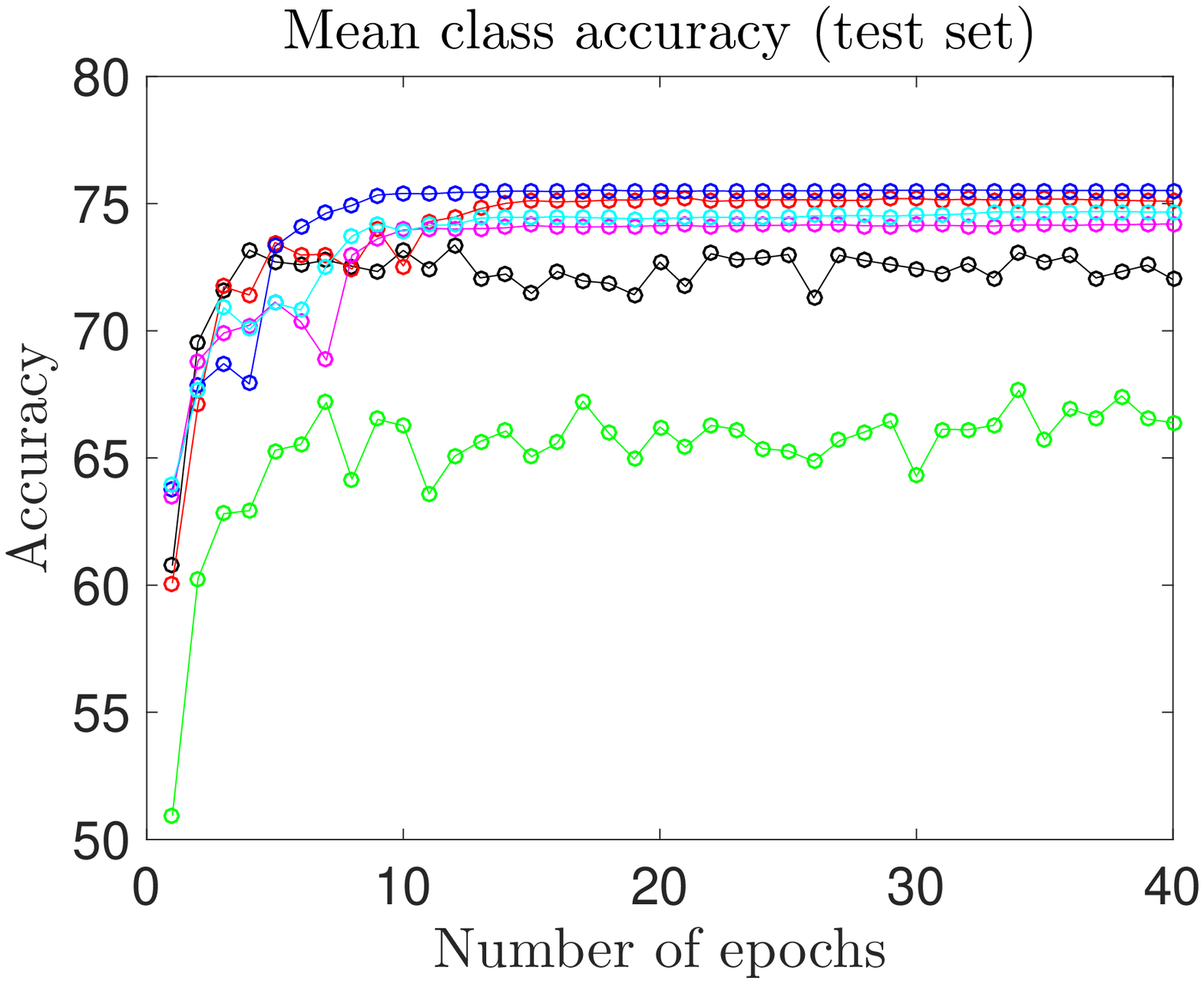}
  \end{subfigure}
  \hfill
  \begin{subfigure}[t]{0.32\textwidth}
    \includegraphics[width=\textwidth]{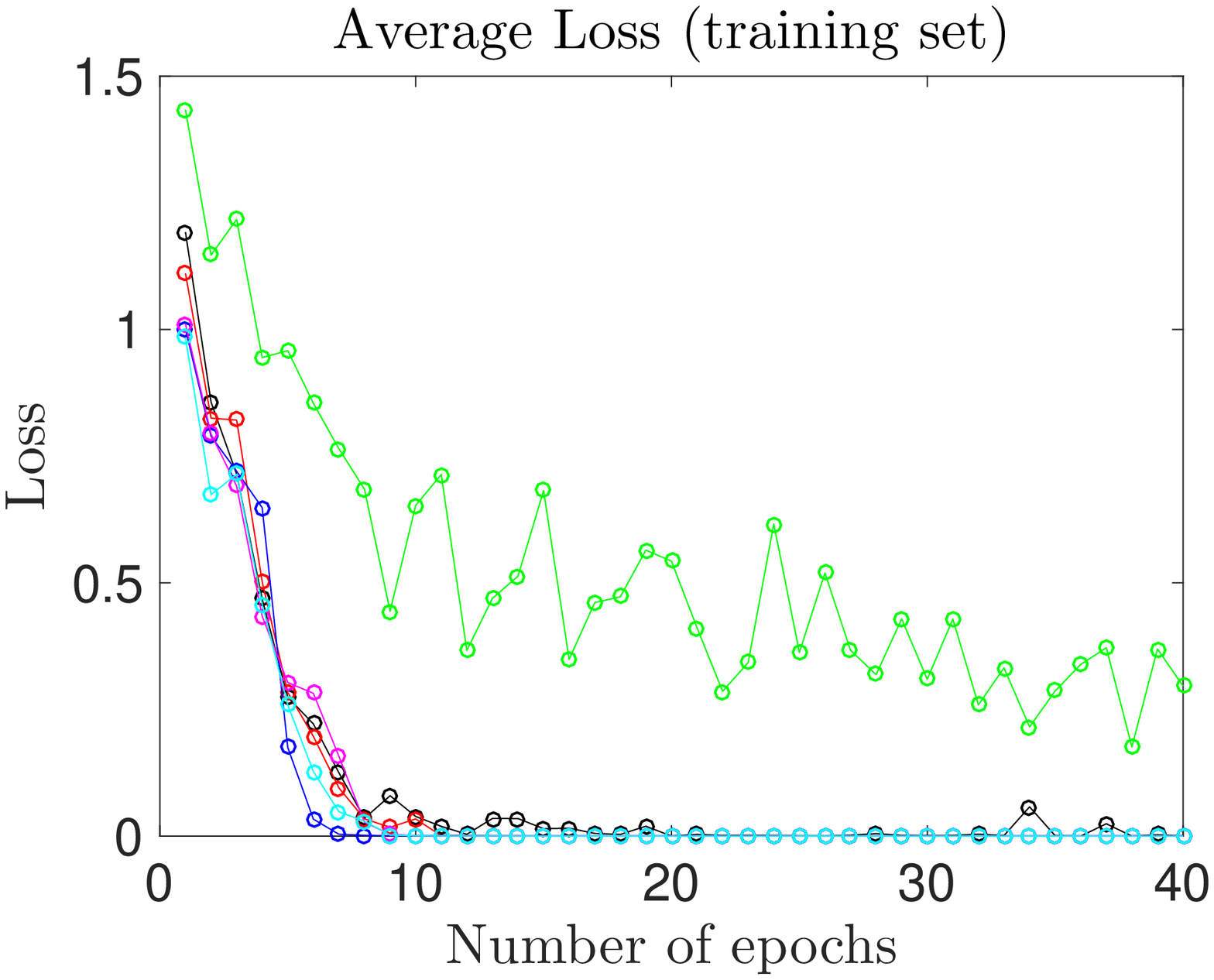}
  \end{subfigure} 
  \hfill
    \begin{subfigure}[t]{0.32\textwidth}
    \includegraphics[width=\textwidth]{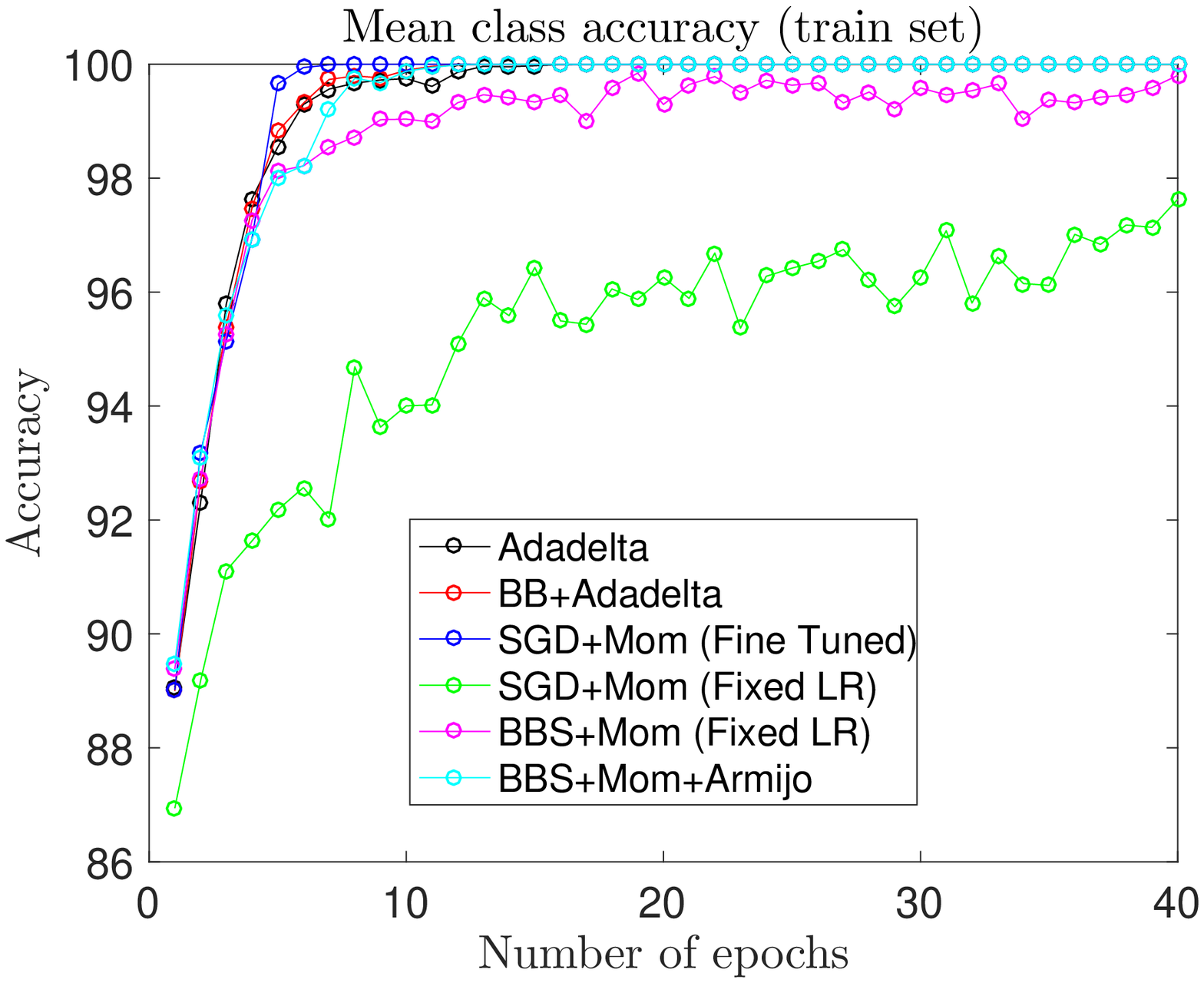}
  \end{subfigure}
  \hfill
  \begin{subfigure}[t]{0.32\textwidth}
    \includegraphics[width=\textwidth]{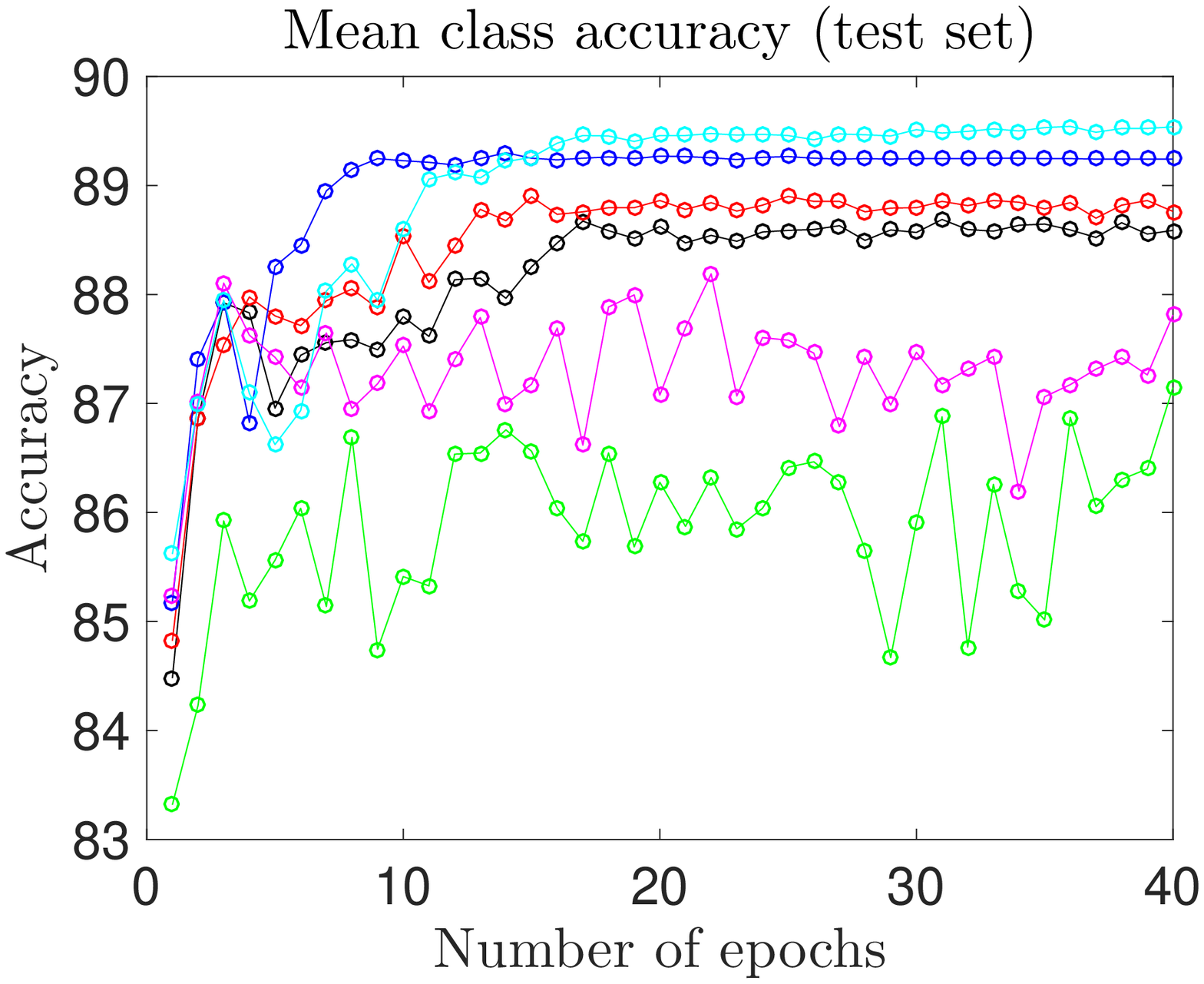}
  \end{subfigure}
  \hfill
  \begin{subfigure}[t]{0.32\textwidth}
    \includegraphics[width=\textwidth]{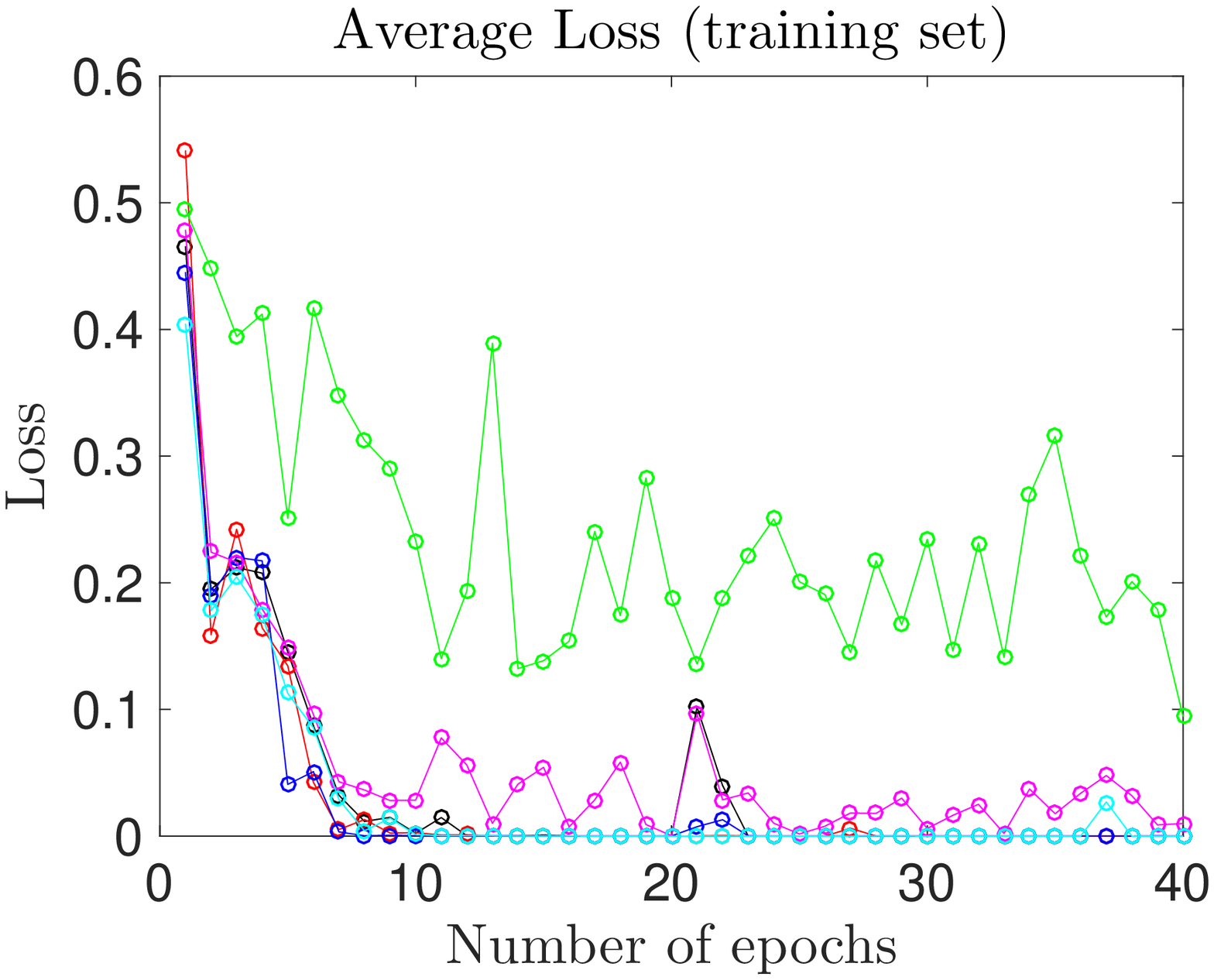}
  \end{subfigure} 
\hfill
   \begin{subfigure}[t]{0.32\textwidth}
    \includegraphics[width=\textwidth]{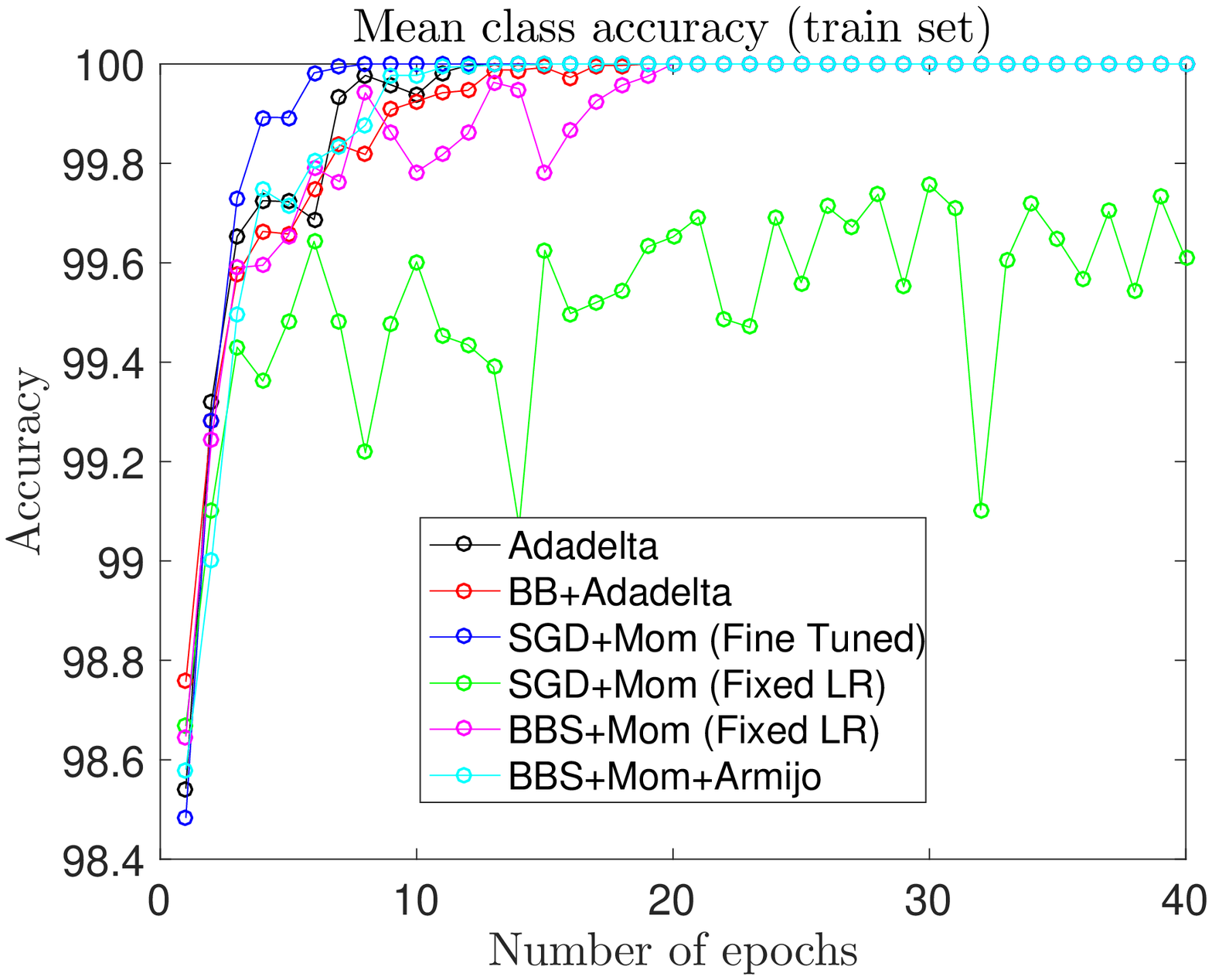}
  \end{subfigure}
  \hfill
  \begin{subfigure}[t]{0.32\textwidth}
    \includegraphics[width=\textwidth]{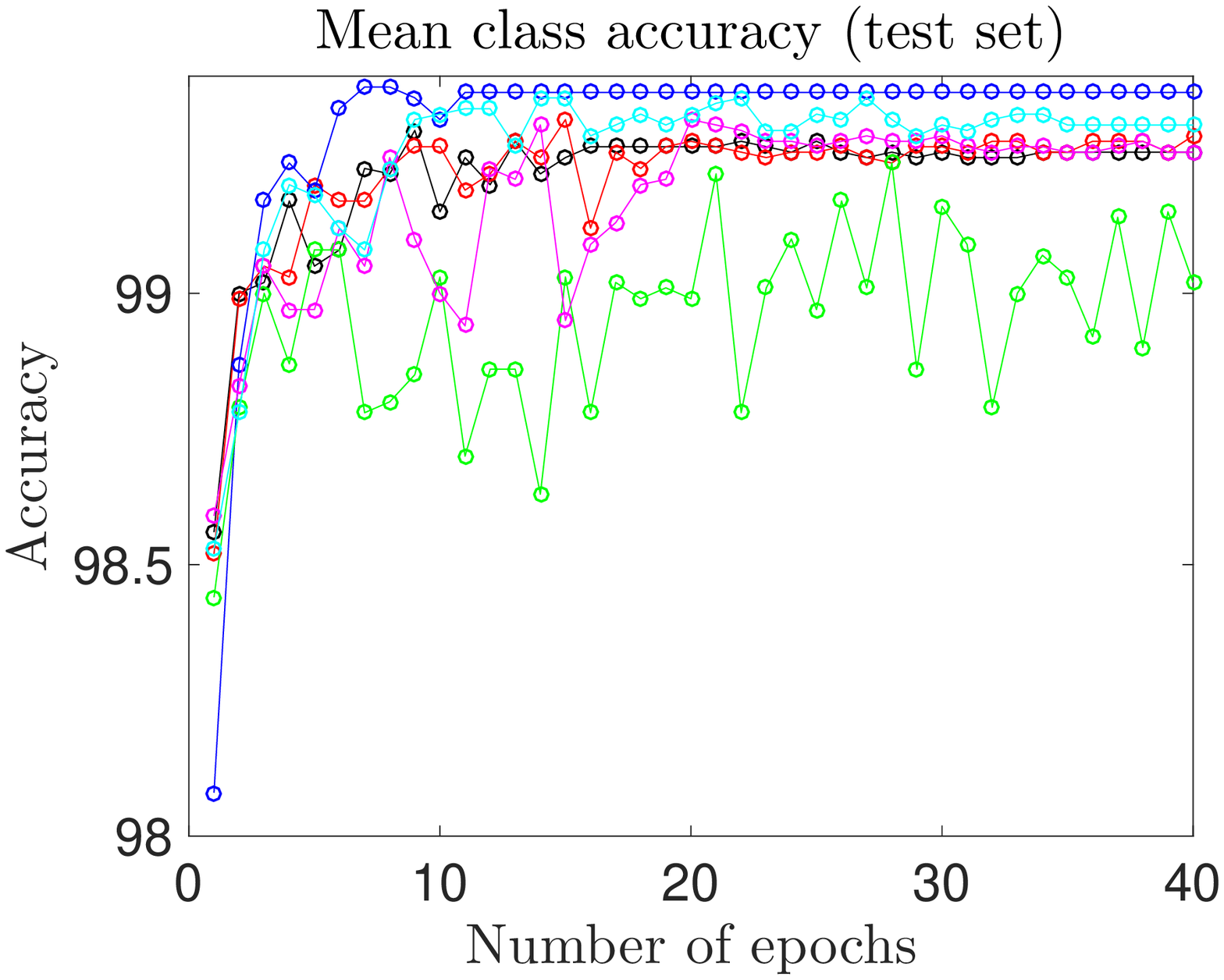}
  \end{subfigure}
  \hfill
  \begin{subfigure}[t]{0.32\textwidth}
    \includegraphics[width=\textwidth]{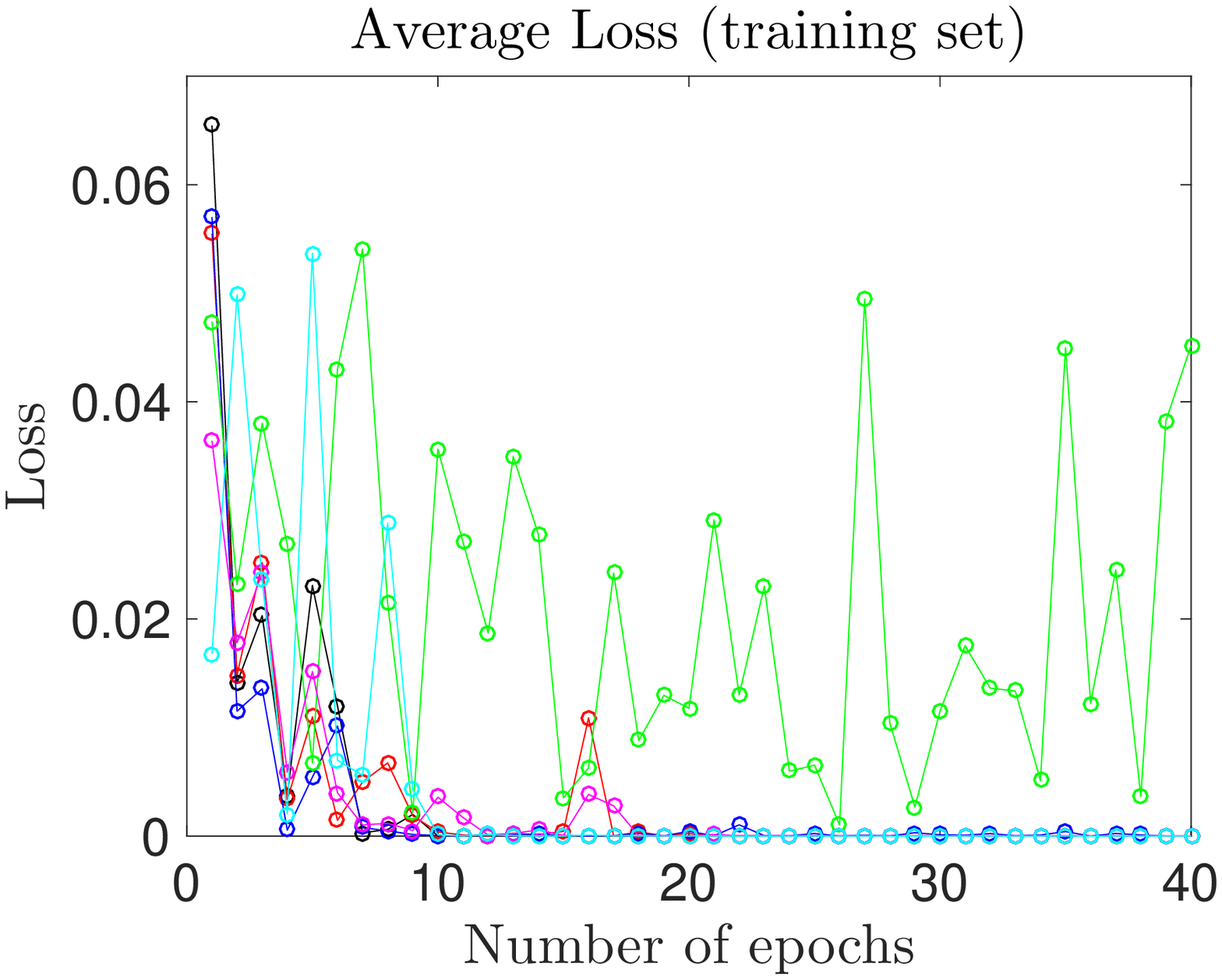}
  \end{subfigure} 
   \caption{ Neural Network Experiments. Top row presents results for CIFAR-10, middle row for SVHN, and bottom row for MNIST. The first column presents classification accuracies on the training set, the middle column presents classification accuracies on the test set, and the last column shows the change in the loss function.}
  \label{fig:neuralnets}  
\end{figure*}

\section{Experiments}
In this section, we present our experimental results. We explore big batch methods with both convex and non-convex (neural network) experiments on large and high-dimensional datasets.

\subsection{Convex Experiments}

For the convex experiments, we test big batch SGD on a binary classification problem with logistic regression and a linear regression problem: 
\begin{align*}
&\min_x \frac{1}{n} \sum_{i=1}^n \log (1 + \exp(-b_i a_i^Tx )), \\
&\min_x \frac{1}{n} \sum_{i=1}^n (a_i^Tx - b_i)^2.
\end{align*}

Figure \ref{fig:convex} presents the results of our convex experiments on three standard real world datasets: IJCNN1 \citep{prokhorov2001ijcnn} and COVERTYPE \citep{blackard1999comparative} for logistic regression, and MILLIONSONG \citep{Bertin-Mahieux2011} for linear regression. As a preprocessing step, we normalize the features for each dataset. We compare deterministic gradient descent (GD) and  SGD with stepsize decay ($\alpha_t = a/(b+t)$) to big batch SGD using a fixed stepsize (BBS+Fixed LR), with backtracking line search (BBS+Armijo) and with the adaptive stepsize \eqref{bb_step} (BBS+BB), as well as the growing batch method described in \citet{friedlander2012hybrid} (denoted as SF; while the authors propose a quasi-Newton method, we adapt their algorithm to a first-order method). We selected stepsize parameters using a comprehensive grid search for all algorithms, except BBS+Armijo and BBS+BB, which require no parameter tuning.

We see that across all three problems, the big batch methods outperform the other algorithms.  We also see that both fully automated methods are always comparable to or better than fixed stepsize methods. 
The automated methods increase the batch size more slowly than BBS+Fixed LR and SF, and thus, these methods can take more steps with smaller batches, leveraging its advantages longer. Further, note that the stepsizes derived by the automated methods are very close to the optimal fixed stepsize rate.

\subsection{Neural Network Experiments}

To demonstrate the versatility of the big batch SGD framework, we also present results on neural network experiments. We compare big batch SGD against SGD with finely tuned stepsize schedules and fixed stepsizes. We also compare with Adadelta~\citep{zeiler2012adadelta}, and combine the big batch method with AdaDelta (BB+AdaDelta) to show that more complex SGD variants can benefit from growing batch sizes. In addition, we had also compared big batch methods with L-BFGS. However, we found L-BFGS to consistently yield poorer generalization error on neural networks, and thus we omitted these results.

We train a convolutional neural network~\citep{lecun1998gradient} (ConvNet) to classify three benchmark image datasets: CIFAR-10~\citep{krizhevsky2009learning}, SVHN~\citep{netzer2011reading}, and MNIST~\citep{lecun1998gradient}. Our ConvNet is composed of $4$ layers. We use $32 \times 32$ pixel images as input. The first layer of the ConvNet contains $16 \times 3 \times 3$, and the second layer contains $256\times 3\times 3$ filters. The third and fourth layers are fully connected ~\citep{lecun1998gradient} with $256$ and $10$ outputs respectively. Each layer except the last one is followed by a ReLu non-linearity ~\citep{krizhevsky2012imagenet} and a max pooling stage~\citep{ranzato2007unsupervised} of size $2 \times 2$. This ConvNet has over 4.3 million weights.

To compare against fine-tuned SGD, we used a comprehensive grid search on the stepsize schedule to identify optimal parameters (up to a factor of 2 accuracy). For CIFAR10, the stepsize starts from $0.5$ and is divided by $2$ every $5$ epochs with $0$ stepsize decay. For SVHN, the stepsize starts from $0.5$ and is divided by $2$ every $5$ epochs with $1e$$-05$ learning rate decay. For MNIST, the learning rate starts from $1$ and is divided by $2$ every $3$ epochs with $0$ stepsize decay. All algorithms use a momentum parameter of $0.9,$ and SGD and AdaDelta use mini-batches of size $128.$ 

Fixed stepsize methods use the default decay rule of the {\em Torch} library: $\alpha_t=\alpha_0/(1+10^{-7}t),$ where $\alpha_0$ was chosen to be the stepsize used in the fine-tuned experiments. We also tune the hyper-parameter $\rho$ in the Adadelta algorithm, and we found $0.9$, $0.9$ and $0.8$ to be best-performing parameters for CIFAR10, SVHN and MNIST respectively.

We plot the accuracy on the train and test set vs the number of epochs (full passes through the dataset) in Figure~\ref{fig:neuralnets}. We notice that the big batch SGD with backtracking performs better than both Adadelta and SGD (Fixed LR) in terms of both train and test error. Big batch SGD even performs comparably to fine tuned SGD but without the trouble of fine tuning. This is interesting because most state-of-the-art deep networks (like AlexNet \citep{krizhevsky2012imagenet}, VGG Net \citep{simonyan2014very}, ResNets \citep{he2016deep}) were trained by their creators using standard SGD with momentum, and training parameters were tuned over long periods of time (sometimes months). Finally, we note that the big batch AdaDelta performs consistently better than plain AdaDelta on both large scale problems (SVHN and CIFAR-10), and performance is nearly identical on the small-scale MNIST problem.



\section{Conclusion}
We analyzed and studied the behavior of alternative SGD methods in which the batch size increases over time.  Unlike classical SGD methods, in which stochastic gradients quickly become swamped with noise, these ``big batch'' methods maintain a nearly constant signal to noise ratio of the approximate gradient.  As a result, big batch methods are able to adaptively adjust batch sizes without user oversight. The proposed automated methods are shown to be empirically comparable or better performing than other standard methods, but without requiring an expert user to choose learning rates and decay parameters.

\subsubsection*{Acknowledgements}

This work was supported by the US Office of Naval Research (N00014-17-1-2078), and the National Science Foundation (CCF-1535902 and IIS-1526234). A. Yadav and D. Jacobs were supported by the Office of the Director
of National Intelligence (ODNI), Intelligence Advanced Research Projects Activity (IARPA), via IARPA R\&D Contract No. 2014-14071600012. The views and conclusions contained herein are those of the authors and should not be interpreted as necessarily representing the official policies or endorsements, either expressed or implied, of the ODNI, IARPA, or
the U.S. Government. The U.S. Government is authorized to reproduce and distribute reprints for Governmental purposes notwithstanding any copyright annotation thereon.

\nocite{goldstein2010high, nesterov2013introductory}
\bibliographystyle{plainnat}
\bibliography{references}

\onecolumn

\begin{center}\huge{Supplementary Material}\end{center}
\appendix

\section{Proof of Lemma \ref{descent_lemma}}
\begin{proof} We know that $- \nabla \ell_\mathcal{B} (x)$ is a descent direction iff the following condition holds:
\begin{equation}
\label{necessary_descent_cond}
\nabla \ell_\mathcal{B} (x)^T\nabla \ell (x) > 0.
\end{equation}
Expanding $ \| \nabla \ell_\mathcal{B} (x) - \nabla \ell (x) \|^2 $ we get
\begin{align}
\| \nabla \ell_\mathcal{B} (x) \|^2 + \| \nabla \ell (x) \|^2 - 2 \nabla \ell_\mathcal{B} (x)^T\nabla \ell (x) &< \| \nabla \ell_\mathcal{B} (x) \|^2, \nonumber \\
\Longrightarrow   - 2 \nabla \ell_\mathcal{B} (x)^T\nabla \ell (x) < - \| \nabla \ell (x) \|^2_2 &\le 0, \nonumber
\end{align}
which is always true for a descent direction \eqref{necessary_descent_cond}. 
\end{proof}

\section{Proof of Theorem \ref{grad_bound}}

\begin{proof}
Let $\bar z = \expect [z] $ be the mean of $z$. Given the current iterate $x$, we assume that the batch $\batch$ is sampled uniformly with replacement from $p$. We then have the following bound:
\begin{align}
 \|\nabla f(x;z)-\nabla \ell(x)\|^2 &\le  2  \|\nabla f(x;z)-\nabla f(x,\bar z)\|^2 + 2\|\nabla f(x,\bar z)-\nabla \ell(x)\|^2  \nonumber  \\
 & \le  2L^2_z \| z- \bar z\|^2 + 2\|\nabla f(x,\bar z)-\nabla \ell(x)\|^2 \nonumber  \\
   & =  2L_z^2 \| z- \bar z\|^2 + 2\| \mathbb{E}_z [\nabla f(x,\bar z)-\nabla f(x, z)] \|^2 \nonumber \\
    & \le  2L_z^2 \| z- \bar z\|^2 +2 \mathbb{E}_z \| \nabla f(x,\bar z)-\nabla f(x, z)\|^2  \nonumber \\
    & \le 2L_z^2 \| z- \bar z\|^2 + 2L_z^2 \mathbb{E}_z  \| \bar z - z\|^2  \nonumber\\
    &= 2L_z^2 \| z- \bar z\|^2 + 2L_z^2 \Tr \Var_z (z), \nonumber
\end{align}
where the first inequality uses the property $\| a + b \|^2 \le 2\| a \|^2 + 2\| b \|^2$, the second and fourth inequalities use Assumption \ref{lipschitz_data}, and the third line uses Jensen's inequality.
This bound is {\em uniform} in $x$.
We then have
\begin{align*}
 \expect_z  \|\nabla f(x;z)-\nabla \ell(x)\|^2 & \le  2L_z^2 \expect_z \| z- \bar z\|^2 + 2L_z^2 \Tr \Var_z (z) \\
&=4L_z^2 \Tr \Var_z (z)
\end{align*}
uniformly for all $x.$ The result follows from the observation that $$\expect_\batch  \|\nabla f_\batch(x)-\nabla \ell(x)\|^2 = \frac{1}{|\batch|} E_z  \|\nabla f(x;z)-\nabla \ell(x)\|^2.$$
\end{proof}

\section{Proof of Lemma \ref{objective_error}}

\begin{proof}
From \eqref{pb_gd} and Assumption \ref{lipschitz_assumption} we get
\begin{equation}
\ell(x_{t+1}) \le \ell (x_t) - \alpha {g_t}^T \nabla \ell(x_t) + \frac{L \alpha^2}{2} \| g_t \|^2. \nonumber
\end{equation}
Taking expectation with respect to the batch $\batch_t$ and conditioning on $x_t$, we get
\begin{align}
\mathbb{E} [\ell(x_{t+1}) - \ell(x\opt)] \le& \ell (x_t) - \ell(x\opt) - \alpha {\mathbb{E}[g_t]}^T \nabla \ell(x_t) + \frac{L \alpha^2}{2} \mathbb{E} \| g_t \|^2 \nonumber \\
=& \ell (x_t) - \ell(x\opt) - \alpha \| \nabla \ell(x_t) \|^2 + \frac{L \alpha^2}{2} (\|\nabla \ell(x_t) \|^2 + \mathbb{E} \| e_t \|^2 + {\mathbb{E} [e_t]}^T \nabla \ell(x_t)) \nonumber \\
=& \ell (x_t) - \ell(x\opt) - \big (\alpha - \frac{L \alpha^2}{2} \big) \| \nabla \ell(x_t) \|^2  + \frac{L \alpha^2}{2}  \mathbb{E} \| e_t \|^2 \nonumber \\
\le& \Big(1 - 2\mu \big (\alpha - \frac{L \alpha^2}{2} \big) \Big) (\ell (x_t) - \ell(x\opt)) + \frac{L \alpha^2}{2}  \mathbb{E} \| e_t \|^2, \nonumber
\end{align}
where the second inequality follows from Assumption \ref{pl_inequality_assumption}. Taking expectation, the result follows.
\end{proof}

\section{Proof of Theorem \ref{strong_rate}}

\begin{proof}
We begin by applying the reverse triangle inequality to \eqref{rewrite_descent_cond} to get 
$$(1-\theta) \expect\| \nabla \ell_\mathcal{B} (x) \| \le \expect \| \nabla \ell (x)\| $$
which applied to \eqref{rewrite_descent_cond} yields
\aln{
\frac{\theta^2}{(1-\theta)^2} \expect \| \nabla \ell (x_t) \|^2 \ge \expect  \| \nabla \ell_\mathcal{B} (x_t) - \nabla \ell (x_t) \|^2 = \expect \|e_t\|^2. \label{swapbound}
}

Now, we apply \eqref{swapbound} to the result in Lemma \ref{objective_error} to get
\aln{
\mathbb{E} [\ell(x_{t+1}) - \ell(x\opt)] \le \expect [\ell (x_t) - \ell(x\opt)] - \big (\alpha - \frac{L \alpha^2\beta }{2}   \big) \expect \| \nabla \ell(x_t) \|^2, \nonumber
}
where $\beta = \frac{\theta^2 + (1-\theta)^2}{(1-\theta)^2} \ge 1.$
Assuming  $\alpha - \frac{L \alpha^2\beta}{2}  \ge 0,$ we can apply 
Assumption \ref{pl_inequality_assumption} to write
\begin{align}
\mathbb{E} [\ell(x_{t+1}) - \ell(x\opt)] \le \Big(1 - 2\mu \big (\alpha - \frac{L \alpha^2\beta}{2} \big) \Big) \expect [\ell (x_t) - \ell(x\opt)],\nonumber
\end{align}
which proves the theorem.  
 Note that $\max_\alpha \{ \alpha - \frac{L \alpha^2\beta}{2}\} = \frac{1}{2L\beta},$ and $\mu \le L.$  It follows that 
 $$ 0 \le \Big(1 - 2\mu \big (\alpha - \frac{L \alpha^2\beta}{2} \big) \Big)  <1.$$
The second result follows immediately.
 \end{proof}

\section{Proof of Theorem \ref{sublinear}}
\begin{proof}
Applying the reverse triangle inequality to \eqref{rewrite_descent_cond} and using Lemma \ref{objective_error} we get, as in Theorem \ref{strong_rate}:
\begin{align}
\label{from_thm2}
\mathbb{E} [\ell(x_{t+1})] \le \expect[\ell (x_t)] - \big (\alpha - \frac{L \alpha^2\beta }{2}   \big) \expect \| \nabla \ell(x_t) \|^2,
\end{align}
where $\beta = \frac{\theta^2 + (1-\theta)^2}{(1-\theta)^2} \ge 1.$ Note that $ \alpha - \frac{L \alpha^2\beta }{2} > 0$ if $\alpha < \frac{2}{L\beta}$. 


From \eqref{pb_gd}, taking norm on both sides and taking expectation, conditioned on all $x_k$, with $k = 0, 1, \cdots, t$, we get
\begin{align}
\mathbb{E} \| x_{t+1} - x\opt \|^2 &= \| x_{t} - x\opt \|^2 - 2\alpha \mathbb{E} \langle x_t - x\opt, \nabla \ell(x_t) + \epsilon_t \rangle + \alpha^2 \mathbb{E} \| \nabla \ell(x_t) + \epsilon_t \|^2 \nonumber \\
&= \| x_{t} - x\opt \|^2 - 2\alpha \langle x_t - x\opt, \nabla \ell(x_t) \rangle + \alpha^2 \| \nabla \ell(x_t) \|^2 + \alpha^2 \mathbb{E} \| \epsilon_t \|^2 \nonumber \\
&\le \| x_{t} - x\opt \|^2 - 2\alpha \langle x_t - x\opt, \nabla \ell(x_t) \rangle + \alpha^2 \| \nabla \ell(x_t) \|^2 + \alpha^2 \frac{\theta^2}{(1-\theta)^2} \| \nabla \ell(x_t) \|^2 \nonumber \\
&= \| x_{t} - x\opt \|^2 - 2\alpha \langle x_t - x\opt, \nabla \ell(x_t) \rangle + \alpha^2 \beta \| \nabla \ell(x_t) \|^2 \nonumber \\
&\le \| x_{t} - x\opt \|^2 - 2\alpha (\ell(x_t) - \ell(x\opt)) + \alpha^2 \beta \| \nabla \ell(x_t) \|^2 \nonumber \\
&\le \| x_{t} - x\opt \|^2 - 2\alpha (\ell(x_t) - \ell(x\opt)) + 2L \alpha^2 \beta (\ell(x_t) - \ell(x\opt)) \nonumber \\
&= \| x_{t} - x\opt \|^2 - (2\alpha - 2L \alpha^2 \beta) (\ell(x_t) - \ell(x\opt)) , \nonumber
\end{align}
where we use the property that $\mathbb{E} [\epsilon_t] = 0$, and the properties $\ell(x) \le \ell(x\opt) + \langle x - x\opt, \nabla \ell(x) \rangle$ (which follows from the convexity of $\ell$) and $\| \nabla \ell(x) \|^2 \le 2L ( \ell(x) - \ell(x\opt))$ (a proof for this identity can be found in \cite{nesterov2013introductory}).


Note that $2\alpha - 2L \alpha^2 \beta > 0$ when $\alpha < \frac{1}{L\beta}$. Taking expectation on all $x$, we get
\begin{align}
\label{to_be_summed}
\mathbb{E} [\ell(x_t) - \ell(x\opt)] \le \frac{1}{2\alpha( 1 - L \alpha \beta)} (\mathbb{E}\| x_{t} - x\opt \|^2 - \mathbb{E} \| x_{t+1} - x\opt \|^2).
\end{align}
Summing \eqref{to_be_summed} over all $t = 0, 1, \cdots, T$, and using the telescoping sum in $\| x_{t} - x\opt \|^2$, we get:
\begin{align}
 \sum_{t = 0}^T \mathbb{E} [\ell(x_t) - \ell(x\opt)] &\le \frac{1}{2\alpha( 1 - L \alpha \beta)} (\mathbb{E}\| x_{0} - x\opt \|^2 - \mathbb{E} \| x_{T+1} - x\opt \|^2)   \nonumber \\
&\le \frac{1}{2\alpha( 1 - L \alpha \beta)} \| x_{0} - x\opt \|^2. \label{summed_weak}
\end{align}
From \eqref{from_thm2} we see that $\mathbb{E} [\ell(x_{t+1}) ] \le \mathbb{E} [ \ell (x_t) ] $ when $\alpha < \frac{2}{L\beta}$. Thus we can write \eqref{summed_weak} as
\begin{align}
\mathbb{E} [\ell(x_T) - \ell(x\opt)] \le \frac{1}{T+1} \sum_{t = 0}^T \mathbb{E} [\ell(x_t) - \ell(x\opt)] \le \frac{ \| x_{0} - x\opt \|^2 }{(2\alpha - 2L \alpha^2 \beta) (T+1)}. \nonumber
\end{align}
Choosing the optimal step size of $\alpha = \frac{1}{2L \beta}$, we get
\begin{align}
\mathbb{E} [\ell(x_T) - \ell(x\opt)]  \le \frac{ 2L\beta \| x_{0} - x\opt \|^2 }{T+1}. \nonumber
\end{align}
\end{proof}

\section{Proof of Theorem \ref{backtracking}}

\begin{proof}
Applying the reverse triangle inequality to \eqref{rewrite_descent_cond} and using Lemma \ref{objective_error} we get, as in Theorem \ref{strong_rate}:
\begin{align}
\label{from_thm2_for4}
\mathbb{E} [\ell(x_{t+1})- \ell(x\opt)] \le \expect[\ell (x_t)- \ell(x\opt)] - \big (\alpha - \frac{L \alpha^2\beta }{2}   \big) \expect \| \nabla \ell(x_t) \|^2,
\end{align}
where $\beta = \frac{\theta^2 + (1-\theta)^2}{(1-\theta)^2} \ge 1.$ 

We will show that the backtracking condition in \eqref{armijo_cond} is satisfied whenever $0 < \alpha_t \le \frac{1}{\beta L}$. Notice that:
\begin{align}
0 < \alpha_t \le \frac{1}{\beta L} \quad \implies \quad - \alpha_t + \frac{L\alpha_t^2 \beta}{2} \le - \frac{\alpha_t}{2}.\nonumber 
\end{align}

Thus, we can rewrite \eqref{from_thm2_for4} as
\begin{align*}
\mathbb{E} [\ell(x_{t+1}) - \ell(x\opt)] &\le \expect [\ell (x_t) - \ell(x\opt) ] - \frac{\alpha_t}{2} \expect \| \nabla \ell(x_t) \|^2 \\
&\le \expect [\ell (x_t) - \ell(x\opt)] - c \alpha_t \expect \| \nabla \ell(x_t) \|^2,
\end{align*}
where $0 < c \le 0.5$. Thus, the backtracking line search condition \eqref{armijo_cond} is satisfied whenever $0 < \alpha_t \le \frac{1}{L \beta}$.

Now we know that either $\alpha_t=\alpha_0$  (the initial stepsize), or $\alpha_t\ge \frac{1}{2\beta L}$, where the stepsize is decreased by a factor of 2 each time the backtracking condition fails. Thus, we can rewrite the above as
\begin{align*}
\mathbb{E} [\ell (x_{t+1}) - \ell(x\opt)] \le \expect [\ell (x_{t}) - \ell(x\opt)] - c \min \Big( \alpha_0, \frac{1}{2\beta L} \Big) \expect \| \nabla \ell (x_t) \|^2.
\end{align*}
Using Assumption \ref{pl_inequality_assumption} we get
\begin{align*}
\expect [\ell (x_{t+1}) - \ell(x\opt)] \le \bigg( 1 - 2c\mu \min \Big( \alpha_0, \frac{1}{2\beta L} \Big) \bigg) \expect[ \ell (x_{t}) - \ell(x\opt) ].
\end{align*}
Assuming we start off the stepsize at a large value such that $\min(\alpha_0, \frac{1}{2\beta L}) = \frac{1}{2 \beta L}$, we can rewrite this as:
\begin{align*}
\expect [\ell (x_{t+1}) - \ell(x\opt)] \le \Big( 1 - \frac{c\mu}{\beta L } \Big)  \expect[ \ell (x_{t}) - \ell(x\opt) ].
\end{align*}
\end{proof}

\section{Derivation of Adaptive Step Size}
Here we present the complete derivation of the adaptive stepsizes presented in Section \ref{sec:adaptive_bb}. Our derivation follows the classical adaptive \citet{barzilai1988two} (BB) method. The BB methods fits a quadratic model to the objective on each iteration, and a stepsize is proposed that is optimal for the local quadratic model \citep{GoldsteinStuderBaraniuk:2014}.
To derive the analog of the BB method for stochastic problems, we consider quadratic approximations of the form $\ell(x) = \expect_\theta f(x, \theta)$, where we define $f(x, \theta) = \frac{\nu}{2} \| x-\theta \|^2$ with $\theta \sim \mathcal{N}(x\opt, \sigma^2I)$. 

We derive the optimal stepsize for this. We can rewrite the quadratic approximation as 
\begin{align*}
\ell (x) = \expect_\theta f(x, \theta) = \frac{\nu}{2} \expect_\theta \| x - \theta \|^2 = \frac{\nu}{2} [ x^Tx - 2x^Tx\opt - \expect (\theta^T \theta) ] = \frac{\nu}{2} \big( \| x - x\opt \|^2 + d\sigma^2 \big),
\end{align*}
since we can write
\begin{align*}
\expect (\theta^T \theta) = \expect \sum_{i=1}^d \theta_i^2 = \sum_{i=1}^d \expect \theta_i^2 = \sum_{i=1}^d (x\opt_i)^2 + \sigma^2 = \| x\opt \|^2 + d\sigma^2.
\end{align*}
Further, notice that:
\begin{align*}
\expect_\theta [\nabla \ell(x) ] &= \expect_\theta [ \nu (x - \theta) ] = \nu (x - x\opt), \quad \text{and} \\
\Tr \Var_\theta [\nabla \ell(x) ] &= \expect_\theta [ \nu^2 (x - \theta)^T(x - \theta) ] - \nu^2 (x - x\opt)^T (x - x\opt) = d \nu^2 \sigma^2.
\end{align*}
Using the quadratic approximation, we can rewrite the update for big batch SGD as follows:
\begin{align*}
x_{t+1} = x_t - \alpha_t  \frac{1}{|\batch|}\sum_{i \in \batch} \nu (x_t - \theta_i) = (1-\nu \alpha_t) x_t + \frac{\nu \alpha_t}{|\batch|} \sum_{i \in\batch} \theta_i = (1-\nu \alpha_t) x_t + \nu \alpha_t x\opt + \frac{\nu \sigma \alpha_t}{|\batch|} \sum_{i \in\batch} \xi_i,
\end{align*}
where we write $\theta_i = x\opt + \sigma \xi_i$ with $\xi_i \sim \mathcal{N}(0,1)$. Thus, the expected value of the function is:
\begin{align*}
\expect [\ell(x_{t+1}) ] &= \expect_\xi \left[\ell \left(  (1-\nu \alpha_t) x_t + \nu \alpha_t x\opt + \frac{\nu \sigma \alpha_t}{|\batch|} \sum_{i \in\batch} \xi_i  \right)\right] \\
&= \frac{\nu}{2} \expect_\xi \left[ \left\| (1-\nu \alpha_t) (x_t - x\opt) + \frac{\nu \sigma \alpha_t}{|\batch|} \sum_{i \in\batch} \xi_i   \right\|^2 + d\sigma^2  \right] \\
&= \frac{\nu}{2} \left( \left\| (1-\nu \alpha_t) (x_t - x\opt) \right\|^2 +   \expect_\xi \left\| \frac{\nu \sigma \alpha_t}{|\batch|} \sum_{i \in\batch} \xi_i   \right\|^2 + d\sigma^2  \right) \\
&= \frac{\nu}{2} \left( \left\| (1-\nu \alpha_t) (x_t - x\opt) \right\|^2 + (1+ \frac{\nu^2 \alpha_t^2}{|\batch|}) d\sigma^2  \right).
\end{align*}
Minimizing $\expect [\ell(x_{t+1}) ]$ w.r.t. $\alpha_t$ we get:
\begin{align}
\alpha_t &= \frac{1}{\nu} \cdot \frac{ \big\| \mathbb{E} [ \nabla \ell_{\batch_t} (x_t) ]  \big\|^2  }{ \big\| \mathbb{E} [ \nabla \ell_{\batch_t} (x_t) ]  \big\|^2 + \frac{1}{|\batch_t|}  \Tr \Var [ \nabla f (x_t) ]  }  \nonumber \\
&= \frac{1}{\nu} \cdot \frac{ \expect \big\|  \nabla \ell_{\batch_t} (x_t)  \big\|^2  - \frac{1}{|\batch_t|}  \Tr \Var [ \nabla f (x_t) ] }{ \expect \big\| \nabla \ell_{\batch_t} (x_t)   \big\|^2  } \nonumber \\
&= \frac{1}{\nu} \cdot \bigg( 1 -  \frac{  \frac{1}{|\batch_t|}  \Tr \Var [ \nabla f (x_t) ] }{ \expect \big\| \nabla \ell_{\batch_t} (x_t)   \big\|^2  }  \bigg) \nonumber \\
&\ge \frac{1 - \theta^2}{\nu} . \nonumber
\end{align}
Thus, the optimal stepsize for big batch SGD is the optimal stepsize for deterministic gradient descent scaled down by at most $1-\theta^2$.

\end{document}